\documentclass[10pt,twocolumn,letterpaper]{article}

\usepackage{cvpr}
\usepackage{rotating}
\usepackage{times}
\usepackage{epsfig}
\usepackage{graphicx}
\usepackage{amsmath}
\usepackage{amssymb}
\usepackage{empheq}
\usepackage{framed, color}
\usepackage{xcolor}
\usepackage{graphics}
\usepackage{graphicx}
\usepackage[]{graphicx} % more modern
\usepackage{epsfig} % less modern
\usepackage{subfigure}
\usepackage{algorithm}
\usepackage{algorithmic}
\usepackage{stmaryrd}
\usepackage[mathscr]{eucal}
\usepackage{lineno}
\usepackage{color}
\usepackage{filecontents}
\usepackage{subfigure}
\usepackage{multirow}
\usepackage{amsmath}
\usepackage{amssymb}
\usepackage{filecontents}
\usepackage{verbatim} % begin{comment} end{comment}
\usepackage{tabularx}
\usepackage{lineno}
\usepackage{setspace}
\usepackage{multirow}
\usepackage{textcomp,booktabs}
\usepackage{caption}
\def\D{{\bf D}}

\def\0{{\bf 0}}
\def\1{{\bf 1}}

\graphicspath{{./figures/}}   %  image

\usepackage[pagebackref=true,breaklinks=true,letterpaper=true,colorlinks,bookmarks=false]{hyperref}

\cvprfinalcopy % *** Uncomment this line for the final submission

\def\cvprPaperID{534} % *** Enter the CVPR Paper ID here

\ifcvprfinal\fi
\begin{document}

%%%%%%%%% TITLE
\title{Recurrent Scene Parsing with Perspective Understanding in the Loop}

\author{
  Shu Kong, \ \ \ \   Charless Fowlkes\\
  Department of Computer Science\\
  University of California, Irvine\\
  Irvine, CA 92697, USA \\
  \texttt{\{skong2, fowlkes\}@ics.uci.edu } \\  \\
  \ [\href{http://www.ics.uci.edu/~skong2/recurrentDepthSeg}{Project Page}]\thanks{Due to size limit of arXiv, all figures included are low-resolution. High-resolution version can be found in the project page. },
[\href{https://github.com/aimerykong/Recurrent-Scene-Parsing-with-Perspective-Understanding-in-the-loop}{
Github}],
[\href{http://www.ics.uci.edu/~skong2/img/depthGatingSeg_poster.pdf}{
Poster}],
[\href{http://www.ics.uci.edu/~skong2/img/depthGatingSeg_slides.pdf}{
Slides}]
}

\maketitle
%\thispagestyle{empty}

%%%%%%%%% ABSTRACT
\begin{abstract}
  Objects may appear at arbitrary scales in perspective images of a scene,
  posing a challenge for recognition systems that process images at a fixed
  resolution.  We propose a depth-aware gating module that adaptively selects
  the pooling field size in a convolutional network architecture according to
  the object scale (inversely proportional to the depth) so that small details
  are preserved for distant objects while larger receptive fields are used for
  those nearby.  The depth gating signal is provided by stereo disparity or
  estimated directly from monocular input.  We integrate this depth-aware
  gating into a recurrent convolutional neural network to perform semantic
  segmentation.  Our recurrent module iteratively refines the segmentation
  results, leveraging the depth and semantic predictions from the previous
  iterations.

  Through extensive experiments on four popular large-scale RGB-D datasets, we
  demonstrate this approach achieves competitive semantic segmentation
  performance with a model which is substantially more compact.  We carry out
  extensive analysis of this architecture including variants that operate on
  monocular RGB but use depth as side-information during training, unsupervised
  gating as a generic attentional mechanism, and multi-resolution gating.
  We find that gated pooling for joint semantic segmentation and depth yields
  state-of-the-art results for quantitative monocular depth estimation.
\end{abstract}

%%%%%%%%% BODY TEXT
\section{Introduction}
\label{sec:intro}

\begin{figure}[t]
\centering
   \includegraphics[width=0.90\linewidth]{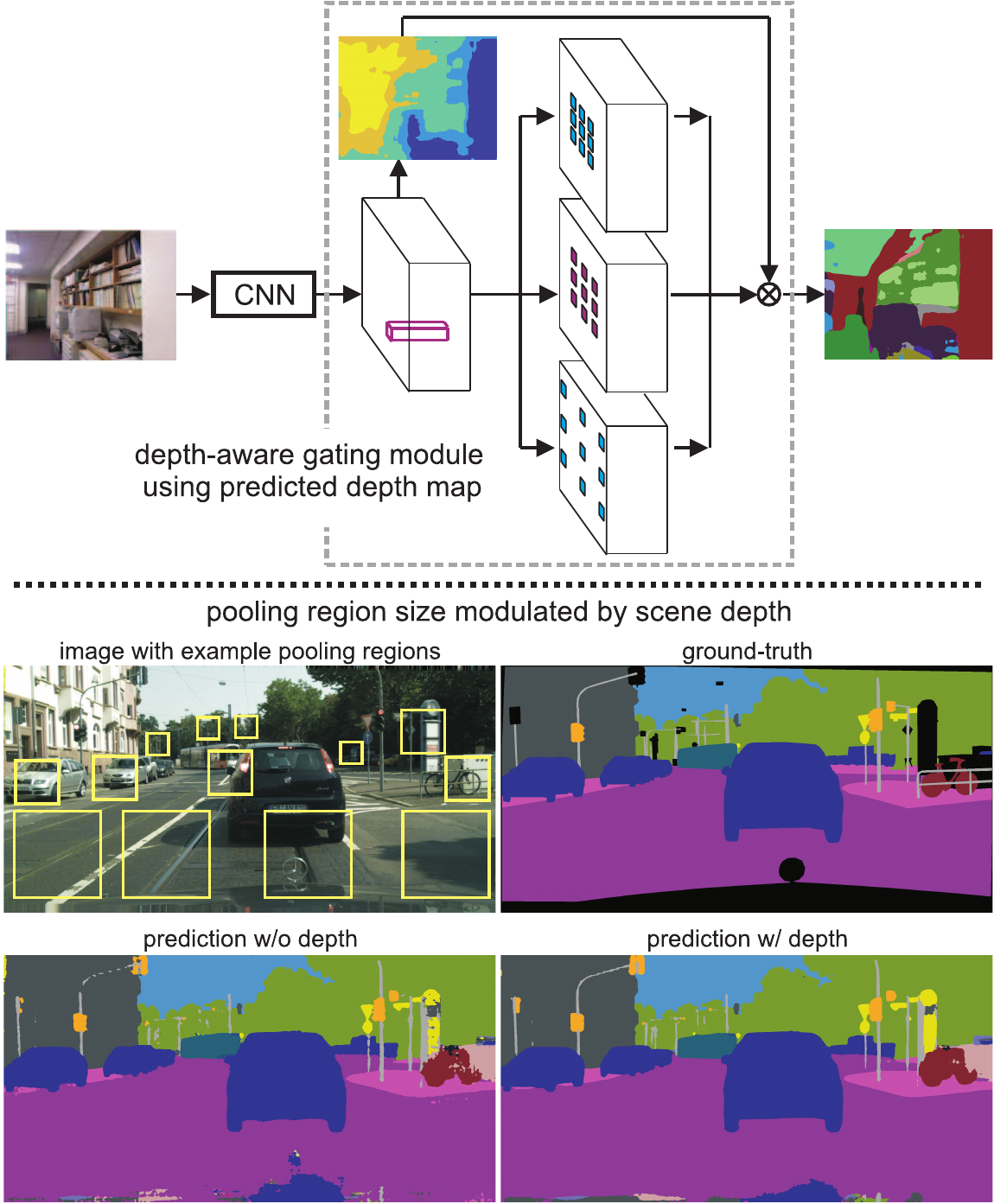}
   \vspace{-3mm}
   \caption{{\bf Upper}: depth-aware gating spatially modulates the selected
   pooling scale using a depth map predicted from monocular input.  In the
   paper, we also evaluate related architectures where scene depth is provided
   directly at test time as a gating signal, and where spatially adaptive
   attentional gating is learned without any depth supervision.
   {\bf Lower}: example ground-truth compared to predictions with and
   without the depth gating module.  Rectangles overlayed on the image indicate
   pooling field sizes which are adapted based on the local depth estimate.
   We quantize the depth map into five discrete scales in our experiments.
   Using depth-gated pooling yields more accurate segment label predictions
   by avoiding pooling across small multiple distant objects while simultaneously
   allowing using sufficiently large pooling fields for nearby objects. }
   \vspace{-12mm}
\label{fig:depthGatingModule}
\end{figure}

An intrinsic challenge of parsing rich scenes is understanding object layout
relative to the camera.  Roughly speaking, the scales of the objects in the
image frame are inversely proportional to the distance to the camera.  Humans
easily recognize objects even when they range over many octaves of spatial
resolution, e.g., the cars near the camera in urban scene can appear a dozen
times larger than those at distance as shown by the lower panel in
Figure~\ref{fig:depthGatingModule}. However, the huge range and arbitrary scale
at which objects appear pose difficulties for machine image understanding.
Although individual local features (e.g., in a deep neural network) can exhibit
some degree of scale-invariance, it is not obvious this invariance covers the
range scale variation that exists in images.

In this paper, we investigate how cues to perspective geometry conveyed by
image content (estimated from stereo disparity, or measured directly via
specialized sensors) might be exploited to improve recognition and scene
understanding.  We focus specifically on the task of semantic segmentation
which seeks to produce per-pixel category labels.

One straightforward approach is to stack the depth map with RGB image as a
four-channel input tensor which can then be processed using standard
architectures.  In practice,
this RGB-D input has not proven successful and sometimes even results in worse
performance~\cite{hazirbas2016fusenet,luo2017unsupervised}.  We conjecture
including depth as a per-pixel input doesn't adequately address
scale-invariance in learning; such models lack an explicit mechanism to
generalize to depths not observed during training and hence still require
training examples with object instances at many different scales to learn a
multiscale appearance model.

Instead, our method takes inspiration from the work of
\cite{ladicky2014pulling}, who propose using depth estimates to rescale local
image patches to a pre-defined canonical depth prior to analysis. For patches
contained within a fronto-parallel surface, this can provide true
depth-invariance over a range of scales (limited by sensor resolution for small
objects) while effectively augmenting the training data available for the
canonical depth.  Rather than rescaling the input image, we propose a depth
gating module that adaptively selects pooling field sizes over higher-level
feature activation layers in a convolutional neural network (CNN).  Adaptive
pooling works with a more abstract notion of scale than standard multiscale
image pyramids which operate on input pixels.  This gating mechanism allows
spatially varying processing over the visual field which can capture context
for semantic segmentation that is not too large or small, but ``just right'',
maintaining details for objects at distance while simultaneously using much
larger receptive fields for objects near the camera. This gating
architecture is trained with a loss that encourages selection of target pooling
scales derived from ``ground-truth'' depth but at test time makes accurate
inferences about scene depth using only monocular cues.

Inspired by studies of human visual processing (e.g.,
\cite{cichy2014resolving}) that suggest dynamic allocation of computation
depending on the task and image content (background clutter, occlusion, object
scale), we propose embedding gated pooling inside a recurrent refinement module
that takes initial estimates of high-level scene semantics as a top-down signal
to reprocess feed-forward representations and refine the final scene
segmentation (similar to the recurrent module proposed
in~\cite{belagiannis2016recurrent} for human pose).  This provides a simple
implementation of ``Biased Competition Theory''~\cite{beck2009top} which allows
top-down feedback to suppress irrelevant stimuli or incorrect interpretations,
an effect we observe qualitatively in our recurrent model near object
boundaries and in cluttered regions with many small objects.

We train this recurrent adaptive pooling CNN architecture end-to-end and
evaluate its performance on several scene parsing datasets. The monocular depth
estimates produced by our gating channel yield state-of-the-art performance on
the NYU-depth-v2 benchmark~\cite{silberman2012indoor}. We also find that using
this gating signal to modulate pooling inside the recurrent refinement
architecture results in improved semantic segmentation performance over
fixed multiresolution pooling.  We also compare to gating models trained
without depth supervision where the gating signal acts as a generic
attentional signal that modulates spatially adaptive pooling. While this works
well, we find that depth supervision results in best performance.  The
resulting system matches state-of-the-art segmentation performance on four
large-scale datasets using a model which, thanks to recurrent computation, is
substantially more compact than many existing approaches.

\section{Related work}

Starting from the ``fully convolutional'' architecture of
~\cite{long2015fully}, there has been a flurry of recent work exploring CNN
architectures for semantic segmentation and other pixel-labeling tasks~\cite{Kong2017recurrent_pixel_embedding}.
The seminal DeepLab~\cite{chen2016deeplab} model modifies the very deep residual
neural network~\cite{he2016deep} for semantic segmentation using dilated or
atrous convolution operators to maintain spatial resolution in high-level
feature maps.  To leverage features conveying finer granularity lower in
the CNN hierarchy, it has proven useful to combine features across
multiple layers (see e.g., FCN~\cite{long2015fully},
LRR~\cite{ghiasi2016laplacian} and RefineNet~\cite{lin2016refinenet}).  To
simultaneously cover larger fields-of-view and incorporate more contextual
information, \cite{zhao2016pyramid} concatenates features pooled over different
scales.
%segmentation using deep neural networks~\cite{wu2016wider,wu2016bridging};
% liu2010single, barinova2008fast

Starting from the work of \cite{hoiem2005geometric, saxena2005learning},
estimating depth from (monocular) scene semantics has been examined in a
variety of indoor and outdoor settings (see e.g., \cite{lee2009geometric}).
Accurate monocular depth estimation using a multiscale deep CNN architecture
was demonstrated by \cite{eigen2014depth} using a geometrically inspired
regression loss.  Follow-on work \cite{eigen2015predicting} showed that depth,
surface orientation and semantic labeling predictions can benefit each other in
a multi-task setting using a shared network model for feature extraction.

The role of perspective geometry and geometric context in object detection was
emphasized by a line of work starting with \cite{hoeim2008putting} and others
(e.g., \cite{bao2011toward}) and has played an increasingly important role,
particularly for scene understanding in urban environments \cite{geiger20143d}.
We were inspired by \cite{ladicky2014pulling}, who showed reliable depth
recovery from image patches (i.e., without vanishing point estimation) and that
the resulting depths could be used to estimate object scale and improve
segmentation in turn.  Chen et al. \cite{chen2016attention} used an attention
gating mechanism to combine predictions from CNN branches run on rescaled
images (multi-resolution), a natural but computationally expensive approach
that we compare experimentally to our proposal (multi-pool).

Finally, there have been a number of proposals to carry out high-level
recognition tasks such as human pose
estimation~\cite{li2016iterative,belagiannis2016recurrent} and semantic
segmentation~\cite{romera2016recurrent}  % arnab2017pixelwise, newell2016stacked
using recurrent or iterative processing.  As pixel-wise labelling tasks are
essentially a structured prediction problem, there has also been a related line
of work that aims to embed unrolled conditional random fields or mean shift into
differentiable CNN architectures to allow for more tractable learning and
inference (e.g., \cite{zheng2015conditional, Kong2017recurrent_pixel_embedding}).

%manually incorporating such structural constraints is impractical and inference
%will be intractable even if such constraints are given.  To deal with this
%dilemma, the recurrent pipeline is expected to learn to discover priors on
%shape, contiguity of region predictions and smoothness of region contours from
%data without any a priori specification~\cite{li2016iterative}.

%\begin{comment}
%\begin{figure}[t]
%\raisebox{-\height}{\vspace{0pt}\includegraphics[width=0.53 \textwidth]{recurrentModule_v2}}\hfill%
%\begin{minipage}[t]{0.43\textwidth}
%%\vspace{-1em}%
%%\vspace{-8mm}%
%\caption{The input to our recurrent module is the concatenation (denoted by
%$\oslash$) of the feature map from an intermediate layer of the feed-forward
%pathway with the prior recurrent prediction. Our recurrent module utilizes
%depth-aware gating which carries out both depth regression and quantized
%prediction. Updated depth predictions at each iteration gate pooling fields
%used for semantic segmentation.  This recurrent update of depth estimation
%increases the flexibility and representation power of our system yielding
%improved segmentation.  We illustrate the prediction prior to, and after two
%recurrent iterations for a particular image. We also visualize the difference
%in predictions between consecutive iterations and note the gains in accuracy at
%each iteration as measured by average intersection-over-union (IoU) benchmark
%performance.}
%\label{fig:recurrentModule}
%\end{minipage}
%%\vspace{-5mm}
%\end{figure}
%\end{comment}

\begin{figure}[t]
\centering
   \includegraphics[width=0.995\linewidth]{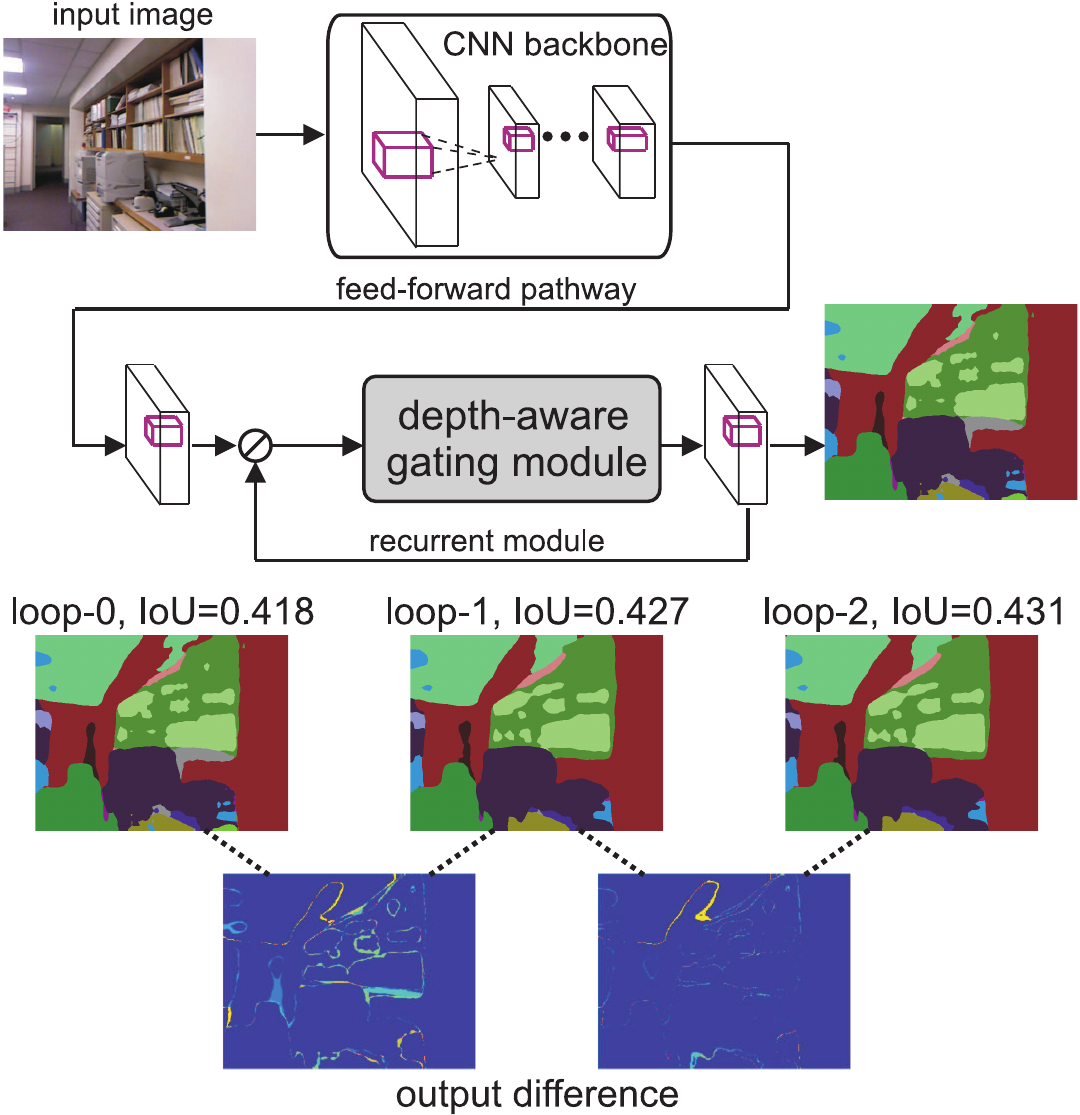}
   %\vspace{-2.5mm}
   \caption{The input to our recurrent module is the concatenation (denoted by
$\oslash$) of the feature map from an intermediate layer of the feed-forward
pathway with the prior recurrent prediction. Our recurrent module utilizes
depth-aware gating which carries out both depth regression and quantized
prediction. Updated depth predictions at each iteration gate pooling fields
used for semantic segmentation.  This recurrent update of depth estimation
increases the flexibility and representation power of our system yielding
improved segmentation.  We illustrate the prediction prior to, and after two
recurrent iterations for a particular image and visualize the difference in
predictions between consecutive iterations which yield small but notable gains
as measured by average intersection-over-union (IoU) benchmark performance.}
%\vspace{-5mm}
\label{fig:recurrentModule}
\end{figure}

\section{Depth-aware Gating Module}
\label{sec:depth}

Our depth-aware gating module utilizes estimated depth at each image location
as a proxy for object scale in order to select the appropriate spatial extent
over which to pool features.  Informally speaking, for a given object category
(e.g., cars) the size of an object in the image is inversely proportional to
the distance from the camera.  Thus, if a region of an image has a larger depth
values, the windows over which features are pooled (pooling field size) should
be smaller in order to avoid pooling responses over many small objects and
capture details needed to precisely segment small objects.  For regions with
small depth values, the same object will appear much larger and the pooling
field size should be scaled up in a covariant manner to capture sufficient
contextual appearance information in the vicinity of the object.

This depth-aware gating can readily utilize depth maps derived from stereo
disparity or specialized time-of-flight sensors.  Such depth maps typically
contain missing data and measurement noise due to oblique view angle,
reflective surface and occlusion boundary.  While these estimates can be
improved using more extensive off-line processing (e.g., \cite{song2015sun}),
in our experiments we use these ``raw'' measurements.  When depth measurements
are not available, the depth-aware gating can instead exploit depth estimated
directly from monocular cues.  The upper panel of
Figure~\ref{fig:depthGatingModule} illustrates the architecture of our
depth-aware gating module using monocular depth predictions derived from the
same front-end feature extractor.

Regardless of the source of the depth map, we quantize the depth into a
discrete set of predicted scales (5 in our experiments). The scale prediction
at each image location is then used to multiplicatively gate between a set of
feature maps computed with corresponding pooling regions and summed to produce
the final feature representation for classification~\cite{kong2016photo,kong2017low}.
In the depth gating
module, we use atrous convolution with different dilation rates to produce the
desired pooling field size on each branch.

When training a monocular depth prediction branch, we quantize the ground-truth
depth and treat it as a five-way classification using a softmax loss. For the
purpose of quantitatively evaluating the accuracy of such monocular depth
prediction, we also train a depth regressor over the input feature of the
module using a simple Euclidean loss for the depth map $\D$ in log-space:
\begin{equation}
\small
\begin{split}
\ell_{depthReg}(\D, \D^*) =& \frac{1}{\vert M \vert}\sum_{(i,j) \in M } \Vert \log(\D_{ij}) - \log(\D_{ij})^*\Vert_2^2,
\nonumber
\end{split}
\end{equation}
where $\D^*$ is the ground-truth depth.  Since our ``ground-truth'' depth may
have missing entries, we only compute the loss over pixels inside a mask $M$
which indicates locations with valid ground-truth depth.  For benchmarking we
convert the log-depth predictions back to depths using an element-wise
exponential.  Although more specific depth-oriented losses have been
explored~\cite{eigen2014depth,eigen2015predicting}, we show in experiment that
this simplistic Euclidean loss on log-depth achieves state-of-the-art monocular
depth estimation when combined with our architecture for semantic segmentation.

In our experiments, we evaluate models based on RGB-D images (where the depth
channel is used for gating) and on RGB images using the monocular depth
estimation branch. We also evaluated a variant which is trained monocularly
(without the depth loss) where the gating can be viewed as a generic attentional
mechanism.  In general, we find that using predicted (monocular) depth
to gate segmentation feature maps yields better performance than models using
the ground-truth depth input.  This is a surprising, but desirable outcome, as
it avoids the need for extra sensor hardware and/or additional computation for
refining depth estimates from multiple video frames (e.g.,
~\cite{song2015sun}).

%We conjecture this is due to two reasons.
%First, the depth prediction branch sharing the network weights increases the representation power and flexibility of the network.
%Second,
%the learned depth predictor outputs object-aware depth maps,
%which gather regions belonging to the same object and can be more helpful in segmentation.
%Figure~\ref{fig:visualization_cityscapes} shows some examples from which we can see some object-aware regions in the predicted depth maps.

%\begin{equation}
%\begin{split}
%H_d(x) =& H_d(W^{w,h}(I,x)) \\
%       =& H_{d/\alpha}(W^{w,h}(\alpha*I, \alpha x)) \\
%       =& H_{d_c}(W^{w,h}(\frac{d}{d_c}*I, \frac{d}{d_c}x)) \\
%\end{split}
%\end{equation}

\section{Recurrent Refinement Module}
\label{sec:recurrent}

It is natural that scene semantics and depth may be helpful in inferring each other.
To achieve this, our recurrent refinement module takes as input feature maps
extracted from a feed-forward CNN model along with current segmentation
predictions available from previous iterations of the recurrent module. These
are concatenated into a single feature map. This allows the recurrent module to
provide an anytime segmentation prediction which can be dynamically refined in
future iterations.  The recurrent refinement module has multiple convolution
layers, each of which is followed by a ReLU and batch normalization layers.  We
also use depth-aware gating in the recurrent module, allowing the refined depth
output to serve as a top-down signal for use in refining the segmentation (as
shown in experiments below).  Figure~\ref{fig:recurrentModule} depicts our
final recurrent architecture using the depth-aware gating module inside.

For a semantic segmentation problem with $K$ semantic classes, we use a $K$-way
softmax classifier on individual pixels to train our network.  Our final
multi-task learning objective function utilizes multiple losses weighted by
hyperparameters:
\begin{equation}
%\small
\begin{split}
\ell = & \sum \limits_{l=0}^{L}(  \lambda_s \ell_{segCls}^l + \lambda_r\ell_{depthReg}^l + \lambda_c\ell_{depthCls}^l), \\
\end{split}
\label{eq:obj}
\end{equation}
where $L$ means we unroll the recurrent module into $L$ loops and $l=0$ denotes
the prediction from the feed-forward pathway.  The three losses
$\ell_{segCls}^l$, $\ell_{depthReg}^l$ and $\ell_{depthCls}^l$ correspond to
the semantic segmentation, depth regression and quantized depth classification
loss at iteration $l$, respectively.  We train our system in a stage-wise
procedure by varying the hyper-parameters $\lambda_s$, $\lambda_r$ and
$\lambda_c$, as detailed in Section~\ref{sec:implementation}, culminating in
end-to-end training using the full objective.  As our primary task is improving
semantic segmentation, in the final training stage we optimize only
$\ell_{segCls}^l$ and drop the depth side-information (setting $\lambda_r=0$
and $\lambda_c=0$).

\section{Implementation}
\label{sec:implementation}
We implement our model with the MatConvNet toolbox~\cite{vedaldi2015matconvnet}
and train using SGD on a single Titan X GPU.  We use the pre-trained ResNet50
and ResNet101 models~\cite{he2016deep} as the backbone of our models\footnote{Code and models are available here: \url{http://www.ics.uci.edu/~skong2/recurrentDepthSeg}.
%, please see project page \url{http://www.ics.uci.edu/~skong2/recurrentDepthSeg}
}.
To increase the output resolution of ResNet, like~\cite{chen2016deeplab}, we
remove the top global $7\times 7$ pooling layer and the last two $2\times2$
pooling layers.  Instead we apply atrous convolution with dilation rate 2 and
4, respectively to maintain a spatial sampling rate which is of $1/8$
resolution to the original image size (rather than $1/32$ resolution if all
pooling layers are kept).  To obtain a final full resolution segmentation
prediction, we simply apply bilinear interpolation on the softmax class scores
to upsample the output by a factor of eight.

We train our models in a stage-wise procedure.  First, we train a feed-forward
baseline model for segmentation.  The feed-forward module is similar to
DeepLab~\cite{chen2016deeplab}, but we add two additional $3\times3$-kernel
layers (without atrous convolution) on top of the ResNet backbone.  Starting
from this baseline, we train depth estimation branch and replace the second
$3\times3$-kernel layer with the depth prediction and depth-aware gating
module.  We train the recurrent refinement module (containing the depth-aware
gating), unrolling one layer at a time, and fine-tune the whole system using
the objective function of Eq.~\ref{eq:obj}.

We augment the training set using random data transforms.  Specifically, we use
random scaling by $s\in [0.5, 2]$, in-plate rotation by degrees in $[-10^\circ,
10^\circ]$, random left-right flip with 0.5 probability, random crop with sizes
around $700\times 700$ divisible by 8, and color jittering.  Note that when
scaling the image by $s$, we also divide the depth values by $s$.  All these
data transforms can be performed in-place with minimal computational cost.

Throughout training, we set batch size to one where the batch is a single input
image (or a crop of a very high-resolution image).  Due to this small batch
size, we freeze the batch normalization in ResNet backbone during training, using
the same constant global moments in both training and testing.  We use the
``poly'' learning rate policy~\cite{chen2016deeplab} with a base learning rate
of $2.5e-4$ scaled as a function of iteration by
$(1-\frac{iter}{maxiter})^{0.9}$.
%The code and trained models will be made
%public\footnote{Code and models are available here: XXXXX.
%, please see project page \url{http://www.ics.uci.edu/~skong2/recurrentDepthSeg}
%}.

\section{Experiments}
\label{sec:exp}
To show the effectiveness of our approach, we carry out comprehensive
experiments on four large-scale RGB-D datasets (introduced below).  We start
with a quantitative evaluation of our monocular depth predictions, which
achieve state-of-the-art performance.
We then compare our complete model with the existing methods for
semantic segmentation on these datasets, followed by ablation experiments to
determine whether our depth-aware gating module improves semantic segmentation,
validate the benefit of our recurrent module, and compare among using
ground-truth depth, predicted depth, and unsupervised attentional gating.
Finally, we show some qualitative results.

\subsection{Datasets and Benchmarks}
For our primary task of semantic segmentation, we use the standard
Intersection-over-Union (IoU) criteria to measure the performance.  We also
report the per-pixel prediction accuracy for the first three datasets to
facilitate comparison to existing approaches.

  \noindent \textbf{NYUD-depth-v2}~\cite{silberman2012indoor} consists of 1,449
  RGB-D indoor scene images of the resolution $640\times 480$ which include
  color and pixel-wise depth obtained by a Kinect sensor.  We use the
  ground-truth segmentation into 40 classes provided
  in~\cite{gupta2013perceptual} and a standard train/test split into 795 and
  654 images respectively.

  \noindent \textbf{SUN-RGBD}~\cite{song2015sun} is an extension of NYUD-depth-v2
  \cite{silberman2012indoor}, containing 5,285 training images and
  5,050 testing images.  It provides pixel labelling masks for 37 classes, and
  depth maps using different depth cameras.  While this dataset provides
  refined depth maps (exploiting depth from the neighborhood video frames), the
  ground-truth depth maps still have significant noisy/mislabled depth
  (examples can be found in our supplemental material).

  \noindent \textbf{Cityscapes}~\cite{cordts2016cityscapes} contains high
  quality pixel-level annotations of images collected in street scenes from 50
  different cities. The training, validation, and test sets contain 2,975, 500,
  and 1,525 images respectively labeled for 19 semantic classes.  The images of
  Cityscapes are of high resolution ($1024\times2048$), which makes training
  challenging due to limited GPU memory.  We randomly crop out sub-images of
  $800\times 800$ resolution for training.

  \noindent \textbf{Stanford-2D-3D}~\cite{armeni2017joint} contains 1,559
  panoramas as well as depth and semantic annotations covering six large-scale
  indoor areas from three different buildings.  We use area 3 and 4 as a
  validation set (489 panoramas) and the remaining four areas for training
  (1,070 panoramas).  The panoramas are very large ($2048\times 4096$) and
  contain black void regions at top and bottom due to the spherical panoramic
  topology.  For the task of semantic segmentation, we rescale them by $0.5$
  and crop out the central two-thirds ($y \in [160, 863]$) resulting in final
  images of size $704\times 2048$-pixels.

{
\setlength{\tabcolsep}{0.35em} % for the horizontal padding
\begin{table}%[th]
\caption{Depth prediction on NYU-depth-v2 dataset. }
\centering
%\vspace{-3mm}
{\small
\begin{tabular}{l | c c c c c c  c c }
\hline
   \vtop{\hbox{\strut Metric} \hbox{\strut $\delta<$ }}
   & \vtop{\hbox{\strut Ladicky}\hbox{\strut \cite{ladicky2014pulling}}}
   & \vtop{\hbox{\strut Liu}\hbox{\strut  \cite{liu2015deep}}}
   & \vtop{\hbox{\strut Eigen}\hbox{\strut  \cite{eigen2014depth}}}
   & \vtop{\hbox{\strut Eigen}\hbox{\strut  \cite{eigen2015predicting}}}
   & \vtop{\hbox{\strut Laina}\hbox{\strut  \cite{laina2016deeper}}}
   & Ours
   & \vtop{\hbox{\strut Ours}\hbox{\strut  -blur}}   \\
\hline
$1.25$               & 0.542     & 0.614 & 0.614 & 0.769 & 0.811 & 0.809 & 0.816\\
$1.25^2$             & 0.829     & 0.883 & 0.888 & 0.950 & 0.953 & 0.945 & 0.950\\
$1.25^3$             & 0.940     & 0.971 & 0.972 & 0.988 & 0.988 & 0.986 & 0.989 \\
\hline
\end{tabular}
}
\label{tab:depthEstComparison}
%\vspace{-2mm}
\end{table}
}
%\shortstack{Ladicky \\ \cite{ladicky2014pulling}}
%\vtop{\hbox{\strut Ladicky}\hbox{\strut \cite{ladicky2014pulling}}}

\subsection{Depth Prediction}

\begin{figure}[t]
\centering
   \includegraphics[width=1.05\linewidth]{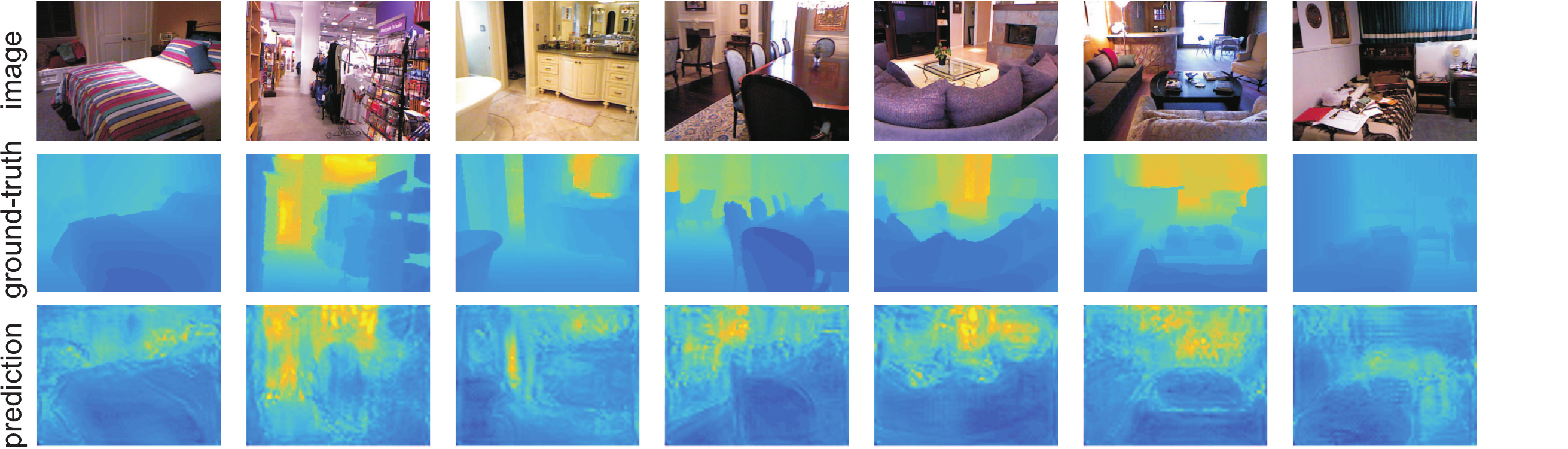}
   %\vspace{-7mm}
   \caption{Examples of monocular depth predictions. First row: the input RGB
   image; second row: ground-truth; third row: our result.  In our
   visualizations, all depth maps use the same fixed (absolute) colormap to
   represent metric depth.}
\label{fig:demo_nyuv2_depth}
%\vspace{-4mm}
\end{figure}

In developing our approach, accurate depth prediction was not the primary goal,
but rather generating a quantized gating signal to select the pooling field
size.  However, to validate our depth prediction, we also trained a depth
regressor over the segmentation backbone and compared the resulting predictions
with previous work.  We evaluated our model on NYU-depth-v2 dataset, on which a
variety of depth prediction methods have been tested.  We report
performance using the standard threshold accuracy metrics, i.e., the percentage
of predicted pixel depths $d_i$ s.t. $\delta = \max(\frac{d_i}{d_i^*},
\frac{d_i^*}{d_i}) < \tau$, evaluated at multiple thresholds $\tau = \{1.25,
1.25^2, 1.25^3\}$.

Table~\ref{tab:depthEstComparison} provides a quantitative comparison of our
predictions with several published methods.  We can see our model trained with
the Euclidean loss on log-depth is quite competitive and achieves significantly
better performance in the $\delta<1.25$ metric. This simplistic loss compares
well to, e.g., \cite{eigen2015predicting} who develop a scale-invariant loss
and use first-order matching term which compares image gradients of the
prediction with the ground-truth, and \cite{laina2016deeper} who develop a set
of sophisticated upsampling layers over a ResNet50 model.

In Figure~\ref{fig:demo_nyuv2_depth}, we visualize our estimated depth maps on
the NYU-depth-v2 dataset\footnote{We also evaluate our depth prediction on
SUN-RGBD dataset, and achieve 0.754, 0.899 and 0.961 by the three threshold
metrics.  As SUN-RGBD is an extension of NYU-depth-v2 dataset, it has similar
data statistics resulting in similar prediction performance.  Examples of depth
prediction on SUN-RGBD dataset can be found in the supplementary material.}.
Visually, we can see our predicted depth maps tend to be noticeably less smooth
than true depth.  Inspired by \cite{eigen2015predicting} who advocate modeling
smoothness in the local prediction, we also apply Gaussian smoothing on our
predicted depth map.  This simple post-process is sufficient to outperform the
state-of-the-art.  We attribute the success of our depth estimator to two
factors.  First, we use a deeper architecture (ResNet50) than that in
\cite{eigen2015predicting} which has generally been shown to improve
performance on a variety vision tasks as reported in literature.  Second, we
train our depth prediction branch jointly with features used for semantic
segmentation.  This is essentially a multi-task problem and the supervision
provided by semantic segmentation may understandably help depth prediction,
explaining why our blurred predictions are as good or better than a similar
ResNet50-based approach which utilized a set of sophisticated upsampling
layers~\cite{laina2016deeper}.

\begin{table*}
%\vspace{-3mm}
\caption{ Performance of semantic segmentation on different datasets.
Results marked by $^\dagger$ are from our trained model models with the released code,
and results marked by $^*$ are evaluated by the dataset server on test set.
Note that we train our models based on ResNet50 architecture on indoor datasets NYU-depth-v2 and SUN-RGBD,
and ResNet101 on the large perspective datasets Cityscapes and Stanford-2D-3D.}
%\vspace{-2mm}
\centering
{\small
\begin{tabular}{l  c c c c c c c c c  c c} % >{\centering}m{1cm}
\hline
                            &  \multicolumn{2}{c} {NYU-depth-v2~\cite{silberman2012indoor}}  &
                            &  \multicolumn{2}{c} {SUN-RGBD~\cite{silberman2012indoor}}     &
                            &  \multicolumn{2}{c} {Stanford-2D-3D~\cite{armeni2017joint}}      &
                            &  Cityscapes~\cite{cordts2016cityscapes}      \\
                            \cmidrule(r){2-3} \cmidrule(r){5-6} \cmidrule(r){8-9} \cmidrule(r){11-11}
                            & IoU & pixel acc. &
                            & IoU & pixel acc. &
                            & IoU & pixel acc. &
                            & IoU    \\
\hline
baseline                    & 0.406   & 0.703 &  & 0.402 & 0.776   & & 0.644 &  0.866  & &  0.738 \\
$\quad$ w/ gt-depth         & 0.413   & 0.708 &  & 0.422 & 0.787   & & 0.730 & 0.897 & &  0.753 \\ % 0.748 tied kernels
$\quad$ w/ pred-depth       & 0.418   & 0.711 &  & 0.423 & 0.789   & & 0.742 & 0.900 & &  0.759 \\
\hline
loop1 w/o depth             & 0.419   & 0.706 &  & 0.432 & 0.793   & & 0.744 & 0.901 & & 0.762 \\
loop1 w/ gt-depth           & 0.425   & 0.711 &  & 0.439 & 0.798   & & 0.747 & 0.902 & & 0.769 \\
loop1 w/ pred-depth         & 0.427   & 0.712 &  & 0.440 & 0.798   & & 0.753 & 0.906 & & 0.772 \\
\hline
loop2                       & 0.431   & 0.713 &  & 0.443 & 0.799   & & 0.760 & 0.908 & & 0.776 \\
loop2 (test-aug)            & 0.445   & 0.721 &  & 0.451 & 0.803   & & 0.765 & 0.910 & & \ \ \ \ \ \ \ \ \ \ \ \ \ \ \ \ 0.791 / $0.782^*$ \\
\hline
%\hline
DeepLab~\cite{chen2016deeplab} & -     & -   & & -      & -         & & 0.698$^\dagger$& 0.880$^\dagger$ & & \ \ \ \ \ \ \ \ \ \ \ \ \ \ \ \  0.704  / $0.704^*$ \\
LRR~\cite{ghiasi2016laplacian} & -   & - &  & -     & -             & & - & - & & \ \ \ \ \ \ \ \ \ \ \ \ \ \ \ \ 0.700 / $0.697^*$ \\
Context~\cite{lin2016efficient} & 0.406 & 0.700   & & 0.423  & 0.784& & - & - & & \ \ \ \ \ \ \ \ \ \ \ \ \ \ \ \ \ \ \ -  \ \ \ \ \  /  $0.716^*$ \\
PSPNet~\cite{zhao2016pyramid}    & -     & -   & & -      & -       & & 0.674$^\dagger$ & 0.876$^\dagger$ & &  \ \ \ \ \ \ \ \ \ \ \ \ \ \ \ \ \ \ \ -  \ \ \ \ \ /  $0.784^*$ \\
RefineNet-Res50~\cite{lin2016refinenet}& 0.438 & -   & & -      & - & & - & - & &  \ \ \ \ \ \ \ \ \ \ \ \ -  \ \ \ \ \ \ /  \ \ \ \ - \\
RefineNet-Res101~\cite{lin2016refinenet}          & 0.447 & -   & & 0.457  & 0.804    & & - & - & &  \ \ \ \ \ \ \ \ \ \ \ \ \ \ \ \ \ \ \ -  \ \ \ \ \  /  $0.736^*$ \\
RefineNet-Res152~\cite{lin2016refinenet}          & 0.465 & 0.736   & & 0.459  & 0.806     & & - & - & &   \ \ \  \ \ \ \ \ \ \ \ \ - \ \ \ \ \ \ /  \ \ \ \ - \\
\hline
\end{tabular}
}
\label{tab:MethodsComparison}
%\vspace{-2mm}
\end{table*}

\subsection{Semantic Segmentation}

To validate the proposed depth-aware gating module and the recurrent refinement
module, we evaluate several variants over our baseline model.
We list the performance details in the first group of rows in
Table~\ref{tab:MethodsComparison}.  The results are consistent across models
trained independently on the four datasets.  Adding depth maps for gating
feature pooling brings noticeable boost in segmentation performance, with
greatest improvements especially on the large-perspective datasets Cityscapes and Stanford-2D-3D.

\begin{figure*}[t]
\centering
   \includegraphics[width=0.999\linewidth]{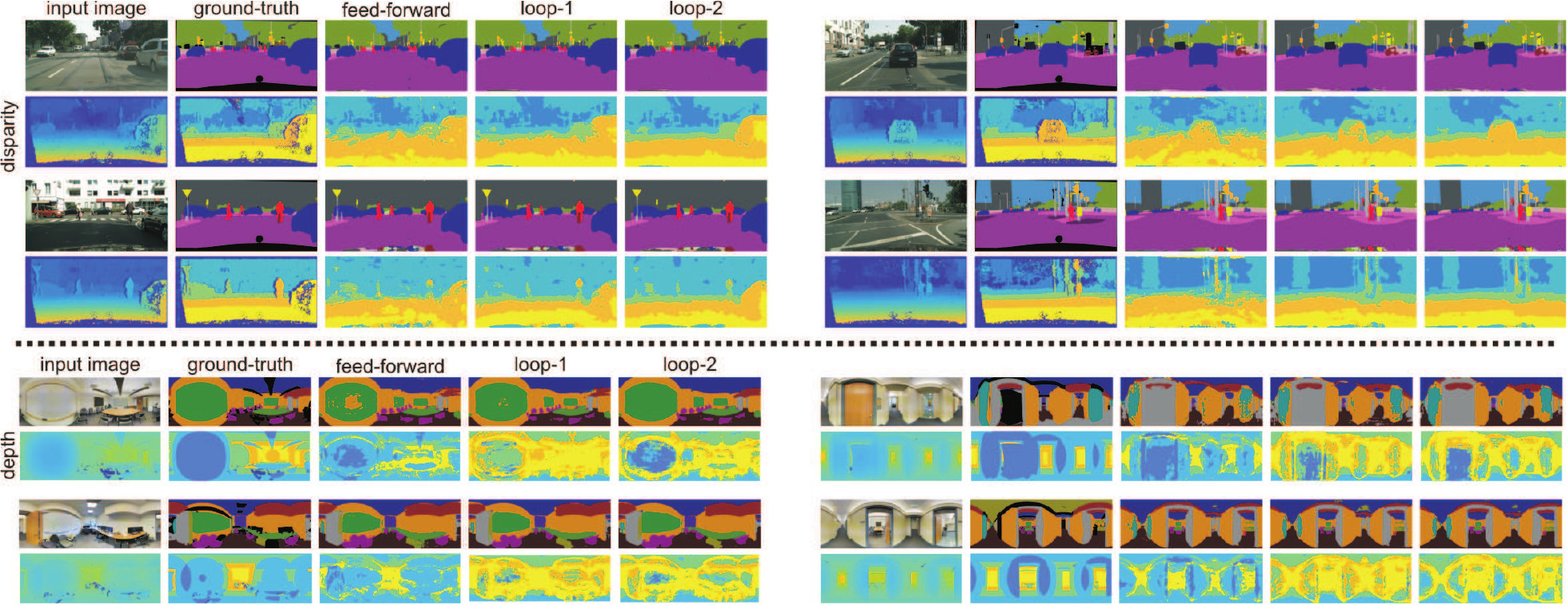} % figure4paper_perspective_demo_lowResolution
   %\vspace{-2mm}
   \caption{Visualization of randomly selected validation images from
   Cityscapes and Stanford-2D-3D with the segmentation output and the predicted
   quantized disparity at each iteration of the recurrent loop.  We depict
   ``ground-truth'' continuous and quantized disparity beneath the input image.
   Our monocular disparity estimate makes predictions for reflective surfaces
   where stereo fails and recurrent iteration further improves estimates,
   particularly for featureless areas such as the pavement.  Note that Cityscapes
   shows disparity while Stanford-2D-3D shows depth so the colormaps are reversed.
   }
\label{fig:visualization_cityscapes}
%\vspace{-3mm}
\end{figure*}

Interestingly, we achieve slightly better performance using the predicted depth
map rather than the provided ground-truth depth.  We attribute this to three
explanations.  Firstly, the predicted depth is smooth without holes or invalid
entries.  When using raw depth, say on Cityscapes and
Stanford-2D-3D\footnote{NYU-depth-v2 and SUN-RGBD datasets provide improved
depth maps without invalid entries.}, we assign equal weight on the missing
entries so that the gating actually averages the information at different
scales.  This average pooling might be harmful in some cases such as a very
small object at a distance.  Secondly, the predicted depth maps show some
object-aware patterns (e.g., car region shown in the visualization in
Figure~\ref{fig:visualization_cityscapes}), which might be helpful for
class-specific segmentation.  Thirdly, the model is trained end-to-end so
co-adaption of the depth prediction and segmentation branches may increase the
overall representation power and flexibility of the whole model, benefiting the
final predictions.

Table~\ref{tab:MethodsComparison} also shows the benefit of the recurrent
refinement module as shown by improved performance from baseline to loop1 and loop2.
Equipped with depth in the recurrent module, the improvement is more notable.
As with the pure feed-forward model, using predicted depth maps in the
recurrent module yields slight gains over the ground-truth depth. We observe
that performance improves using a depth 2 unrolling (third group of rows
in Table~\ref{tab:MethodsComparison}) but saturates/converges after two
iterations.

In comparing with state-of-the-art methods, we follow common practice of
augmenting images at test time by running the model on flipped and rescaled
variants and average the class scores to produce the final segmentation output
(compare loop2 and loop2 (test-aug)).  We can see our model performs on par or
better than recently published results listed in
Table~\ref{tab:MethodsComparison}.

Note that for NYU-depth-v2 and SUN-RGBD, our backbone architecture is ResNet50,
whereas RefineNet reports the results using a much deeper models (ResNet101 and
ResNet152) which typically outperform shallower networks in vision tasks.
For the Cityscapes, we also submitted our final result for held-out
benchmark images which were evaluated by the Cityscapes benchmark server.
Our model achieves IoU $0.782$, on par with the best published result, IoU
$0.784$, by PSPNet.\footnote{We compare to performance using train only
rather than train+val which improved PSPNet performance to $0.813$.}
We did not perform any extensive performance tuning and only utilized the
fine-annotation training images for training (without the twenty thousand
coarse-annotation images and the validation set). We also didn't utilize any
post-processing (such as the widely used fully-connected
CRF~\cite{koltun2011efficient} which typical yields additional performance
increments).

One key advantage of recurrent refinement is that it allows richer computation
(and better performance) without additional model parameters.  Our ResNet50
model (used on the NYU-depth-v2 dataset) is relatively compact (221MB) compared
to RefineNet-Res101 which achieves similar performance but is nearly double the
size (426MB). Our model architecture is similar to DeepLab which also adopts
pyramid atrous convolution at multiple scales of inputs (but simply averages
output feature maps without any depth-guided adaptive pooling).  However, the
final DeepLab model utilizes an ensemble which yields a much larger model
(530MB).  PSPNet concatenates the intermediate features into 4,096 dimension
before classification while our model operates on small 512-dimension feature
maps.

\begin{figure}[t]
\centering
   \includegraphics[width=0.9985\linewidth]{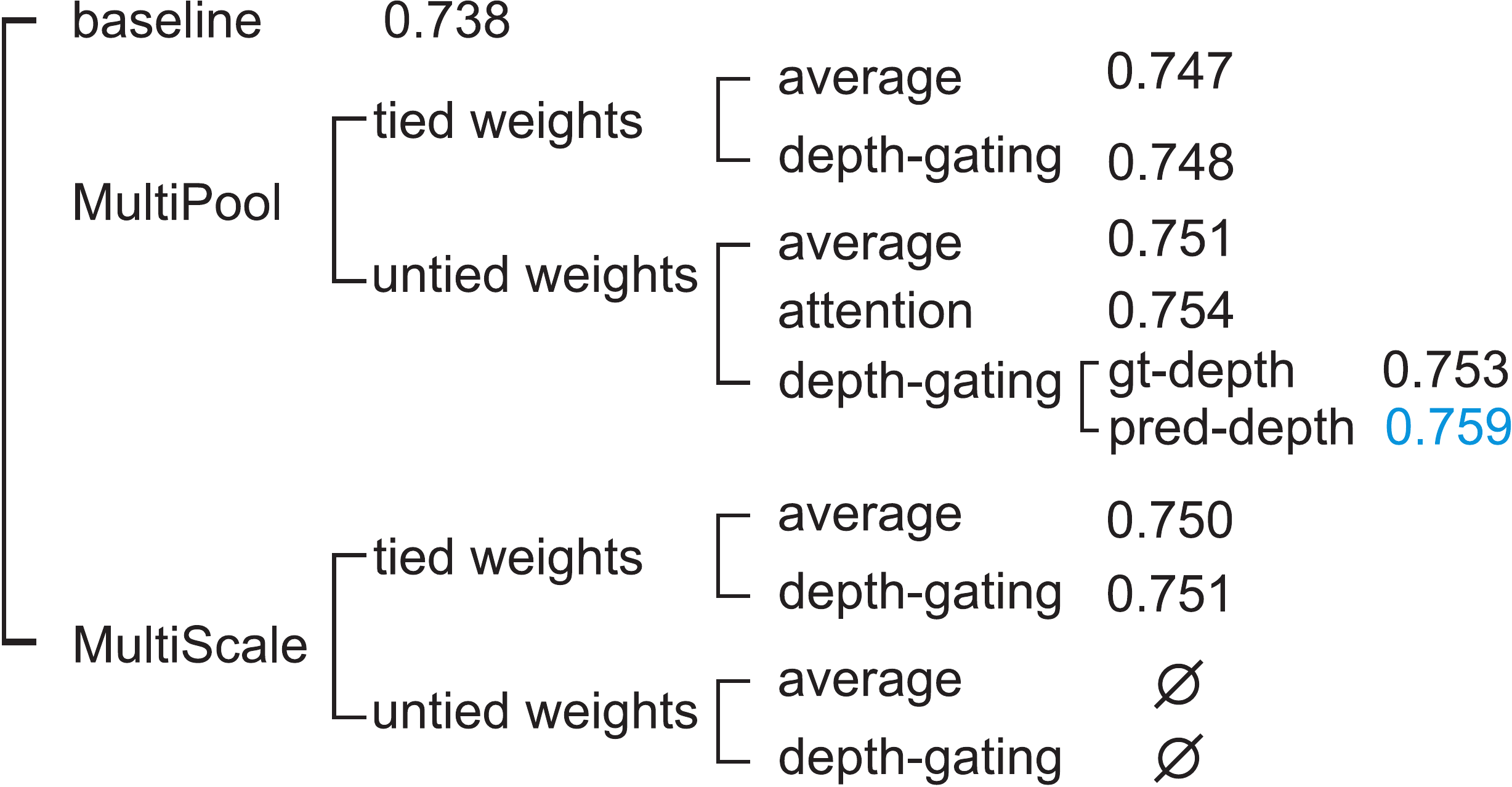}
 %  \vspace{-2mm}
   \caption{Performance comparisons across gating architectures including tied vs untied
   parameters across different branches, averaging vs gating branch predictions, using
   monocular predicted vs ground-truth depth for the gating signal, gating pooling region
   size (MultiPool) or rescaling input image (MultiScale), and gating without depth
   supervision during training (attention).}
\label{fig:ablationStudy}
%\vspace{-5mm}
\end{figure}

\begin{figure}[t]
\centering
   \includegraphics[width=0.9985\linewidth]{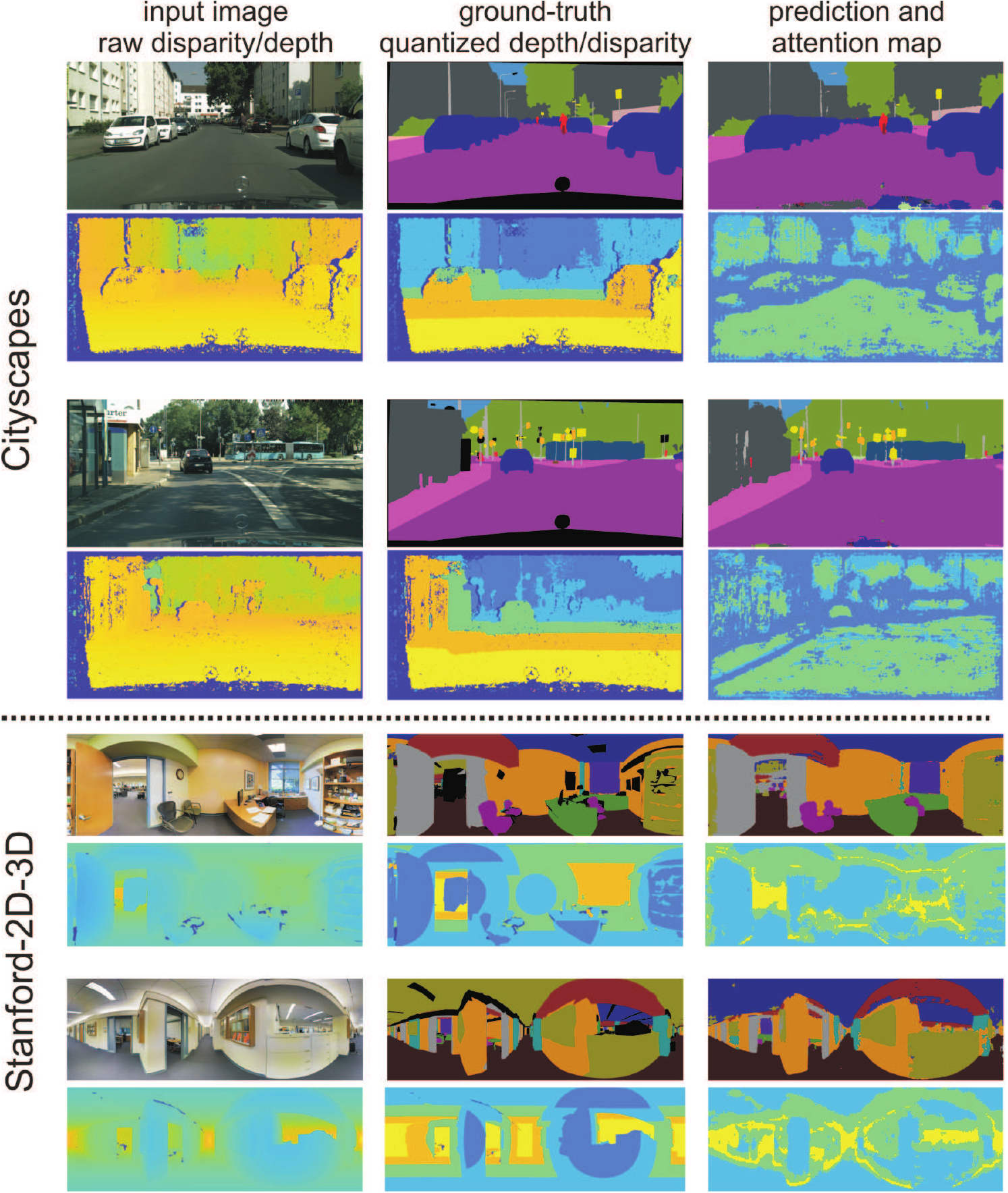}
 %  \vspace{-3mm}
   \caption{Visualization of the attention maps on random images from
   Cityscapes and Stanford-2D-3D.  The raw disparity/depth maps and the
   quantized versions are also shown for reference.
   Though we train the attention branch with randomly initialized weights, we
   can see that the learned attention maps capture some depth information as
   well as encoding distance to object boundaries.
   %\TODO{change caxis limits on stanford data so that more is visible in the raw ground-truth image}
   }
\label{fig:demo_attention}
%\vspace{-2mm}
\end{figure}

\begin{figure}[t]
\centering
   \includegraphics[width=0.99\linewidth]{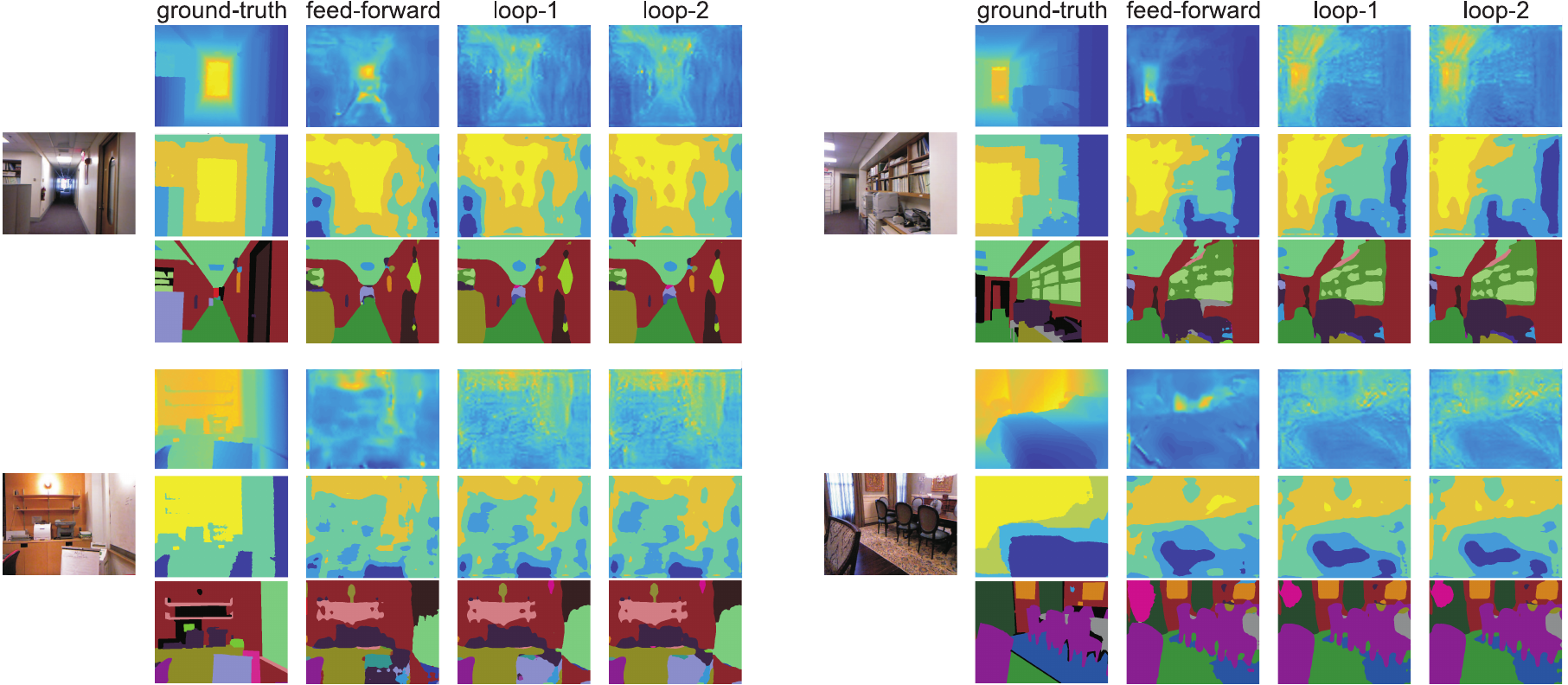}
   %\vspace{-3mm}
   \caption{Visualization of the output on NYU-depth-v2.  We show four randomly
   selected testing images with ground-truth and predicted disparity (first row),
   quantized disparity (second row) and segmentation (third row) at each iteration
   of the recurrent computation.
   }
\label{fig:visualization_nyuv2}
%\vspace{-4mm}
\end{figure}

\subsection{Analysis of Gating Architectures Alternatives}

We  discuss the important question of whether depth-aware gating
is really responsible for improved performance over baseline, or if gains are
simply attributable to training a larger, richer architecture. We also contrast
our approach to a number of related proposals in the literature.  We summarize
our experiments exploring these alternatives in Figure~\ref{fig:ablationStudy}
(more details can be found in supplementary material).

We use the term {\em MultiPool} to denote the family of models (like our
proposed model) which process the input image at a single fixed scale, but
perform pooling at multiple convolutional dilate rate at high level layers.
For a multi-pool architecture, we may choose to learn independent {\em untied}
weights across the scale-specific branches or use the same {\em tied} weights.
As an alternative to our gating function, which selects a spatially varying
weighted combination of the scale-specific branches, we can simply {\em
average} the branches (identical at all spatial locations).

We can contrast MultiPool with the {\em MultiScale} approach, which combines
representations or predictions from multiple branches where each branch is
applied to a scaled version of the input image\footnote{The roots of this
idea can be traced back to early work on scale-space for edge detection
(see, e.g. \cite{bergholm1987edge, lindeberg1998feature})}.  Many have adopted
this strategy as a test time heuristic to boost performance by simply running
the same model (tied) on different scaled versions of the input and then
averaging the predictions. Others, such as DeepLab~\cite{chen2016deeplab},
train multiple (untied) models and use the average ensemble output.
%\TODO{add old scale space citations}

In practice, we found that both MultiPool and MultiScale architectures
outperform baseline and achieve similar performance.  While MultiScale
processing is conceptually appealing, it has a substantial computational
overhead relative to MultiPool processing (where early computation is shared
among branches). As a result, it was not feasible to train untied MultiScale
models end-to-end on a single GPU memory constraints. As a result, we found
that the untied, depth-gated model performed the best (and was adopted
in our final approach).

Finally, we explored use of the gated pooling where the gating was trained without
the depth loss. We refer to this as an {\em attention} model after the work of
\cite{chen2016attention}.  The attention model achieves surprisingly good
performance, even outperform gating using ground-truth depth.  We show the
learned attention map in Figure~\ref{fig:demo_attention}, which behaves quite
differently from depth gating. Instead, the gating signal appears to encode the
distance from object boundaries. We hypothesize this selection mechanism serves
to avoid pooling features across different semantic segments while still
utilizing large pooling regions within each region. Our fine-tuned model using
predicted depth-gating (instead of ground-truth depth) likely benefits from
this adaption.

\subsection{Qualitative Results}

In Figures~\ref{fig:visualization_nyuv2} and \ref{fig:visualization_cityscapes},
we depict several randomly selected examples from the test set of
NYU-depth-v2, Cityscapes and Stanford-2D-3D.  We visualize both the
segmentation results and the depth maps updated across multiple recurrent
iterations.  Interestingly, the depth maps on Cityscapes and Stanford-2D-3D
change more noticeably than those on NYU-depth-v2 dataset.
In Cityscapes, regions in the predicted depth map corresponding to objects,
such as the car, are grouped together and disparity estimates on texture-less
regions such as the street surface improve across iterations, while in
Stanford-2D-3D, depth estimate for adaptation suggests that the recurrent
module is performing coarse-to-fine segmentation (where later iterations
shift towards a smaller pooling regions as semantic confidence increases).

Gains for the NYU-depth-v2 data are less apparent.  We conjecture
this is because images in NYU-depth-v2 are more varied in overall layout and
often have less texture and fewer objects from which the model can infer
semantics and subsequently depth.  In all datasets, we can see that our model
is able to exploit recurrence to correct misclassified regions/pixels ``in the
loop'', visually demonstrating the effectiveness of the recurrent refinement
module.

\section{Conclusion and Discussion}
In this paper, we have proposed a depth-aware gating module that uses depth
estimates to adaptively modify the pooling field size at a high level layer of
neural network for better segmentation performance.  The adaptive pooling can
use large pooling fields to include more contextual information for labeling
large nearby objects, while maintaining fine-scale detail for objects further
from the camera. While our model can utilize stereo disparity directly, we find
that using such data to train a depth predictor which is subsequently used for
adaptation at test-time in place of stereo ultimately yields better
performance.  We also demonstrate the utility of performing recurrent
refinement which yields improved prediction accuracy for semantic
segmentation without adding additional model parameters.

We envision that the recurrent refinement module can capture object shape
priors, contour smoothness and region continuity. However, our current approach
converges after a few iterations and performance saturates. This leaves open
future work in exploring other training objectives that might push the
recurrent computation towards producing more varied outputs. This might be
further enriched in the setting of video where the recurrent component could be
extended to incorporate memory of previous frames.

\section*{Acknowledgement}
This project is supported by NSF grants
IIS-1618806, IIS-1253538, DBI-1262547 and a hardware donation
from NVIDIA.

%\begin{table}%[th]
%\small
%\centering
%\caption{Depth Prediction on SUN-RGBD dataset.}
%%\vspace{-3mm}
%\begin{tabular}{l | c  }
%\hline
%                            & Ours    \\
%\hline
%$\delta<1.25$               & 0.754   \\
%$\delta<1.25^2$             & 0.899   \\
%$\delta<1.25^3$             & 0.961   \\
%\hline
%\end{tabular}
%\label{tab:SUNRGBD_depth}
%\end{table}

%\clearpage\mbox{}Page \thepage\ of the manuscript.
%\clearpage\mbox{}Page \thepage\ of the manuscript.

{\small
\bibliography{egbib}

\begin{thebibliography}{10}\itemsep=-1pt

\bibitem{armeni2017joint}
I.~Armeni, S.~Sax, A.~R. Zamir, and S.~Savarese.
\newblock Joint 2d-3d-semantic data for indoor scene understanding.
\newblock {\em arXiv preprint arXiv:1702.01105}, 2017.

\bibitem{bao2011toward}
S.~Y. Bao, M.~Sun, and S.~Savarese.
\newblock Toward coherent object detection and scene layout understanding.
\newblock {\em Image and Vision Computing}, 29(9):569--579, 2011.

\bibitem{beck2009top}
D.~M. Beck and S.~Kastner.
\newblock Top-down and bottom-up mechanisms in biasing competition in the human
  brain.
\newblock {\em Vision research}, 49(10):1154--1165, 2009.

\bibitem{belagiannis2016recurrent}
V.~Belagiannis and A.~Zisserman.
\newblock Recurrent human pose estimation.
\newblock {\em arXiv:1605.02914}, 2016.

\bibitem{bergholm1987edge}
F.~Bergholm.
\newblock Edge focusing.
\newblock {\em IEEE Transactions on Pattern Analysis and Machine Intelligence},
  (6):726--741, 1987.

\bibitem{chen2016deeplab}
L.-C. Chen, G.~Papandreou, I.~Kokkinos, K.~Murphy, and A.~L. Yuille.
\newblock Deeplab: Semantic image segmentation with deep convolutional nets,
  atrous convolution, and fully connected crfs.
\newblock {\em arXiv preprint arXiv:1606.00915}, 2016.

\bibitem{chen2016attention}
L.-C. Chen, Y.~Yang, J.~Wang, W.~Xu, and A.~L. Yuille.
\newblock Attention to scale: Scale-aware semantic image segmentation.
\newblock In {\em Proceedings of the IEEE Conference on Computer Vision and
  Pattern Recognition}, pages 3640--3649, 2016.

\bibitem{cichy2014resolving}
R.~M. Cichy, D.~Pantazis, and A.~Oliva.
\newblock Resolving human object recognition in space and time.
\newblock {\em Nature neuroscience}, 17(3):455--462, 2014.

\bibitem{cordts2016cityscapes}
M.~Cordts, M.~Omran, S.~Ramos, T.~Rehfeld, M.~Enzweiler, R.~Benenson,
  U.~Franke, S.~Roth, and B.~Schiele.
\newblock The cityscapes dataset for semantic urban scene understanding.
\newblock In {\em Proceedings of the IEEE Conference on Computer Vision and
  Pattern Recognition}, 2016.

\bibitem{eigen2015predicting}
D.~Eigen and R.~Fergus.
\newblock Predicting depth, surface normals and semantic labels with a common
  multi-scale convolutional architecture.
\newblock In {\em Proceedings of the IEEE International Conference on Computer
  Vision}, pages 2650--2658, 2015.

\bibitem{eigen2014depth}
D.~Eigen, C.~Puhrsch, and R.~Fergus.
\newblock Depth map prediction from a single image using a multi-scale deep
  network.
\newblock In {\em NIPS}, 2014.

\bibitem{geiger20143d}
A.~Geiger, M.~Lauer, C.~Wojek, C.~Stiller, and R.~Urtasun.
\newblock 3d traffic scene understanding from movable platforms.
\newblock {\em IEEE transactions on pattern analysis and machine intelligence},
  36(5):1012--1025, 2014.

\bibitem{ghiasi2016laplacian}
G.~Ghiasi and C.~C. Fowlkes.
\newblock Laplacian pyramid reconstruction and refinement for semantic
  segmentation.
\newblock In {\em ECCV}, 2016.

\bibitem{gupta2013perceptual}
S.~Gupta, P.~Arbelaez, and J.~Malik.
\newblock Perceptual organization and recognition of indoor scenes from rgb-d
  images.
\newblock In {\em Proceedings of the IEEE Conference on Computer Vision and
  Pattern Recognition}, 2013.

\bibitem{hazirbas2016fusenet}
C.~Hazirbas, L.~Ma, C.~Domokos, and D.~Cremers.
\newblock Fusenet: Incorporating depth into semantic segmentation via
  fusion-based cnn architecture.
\newblock In {\em ACCV}, 2016.

\bibitem{he2016deep}
K.~He, X.~Zhang, S.~Ren, and J.~Sun.
\newblock Deep residual learning for image recognition.
\newblock In {\em Proceedings of the IEEE Conference on Computer Vision and
  Pattern Recognition}, 2016.

\bibitem{hoiem2005geometric}
D.~Hoiem, A.~A. Efros, and M.~Hebert.
\newblock Geometric context from a single image.
\newblock In {\em Proceedings of the IEEE International Conference on Computer
  Vision}, 2005.

\bibitem{hoeim2008putting}
D.~Hoiem, A.~A. Efros, and M.~Hebert.
\newblock Putting objects in perspective.
\newblock {\em International Journal of Computer Vision}, 80(1):3--15, 2008.

\bibitem{kong2017low}
S.~Kong and C.~Fowlkes.
\newblock Low-rank bilinear pooling for fine-grained classification.
\newblock In {\em 2017 IEEE Conference on Computer Vision and Pattern
  Recognition (CVPR)}, pages 7025--7034. IEEE, 2017.

\bibitem{Kong2017recurrent_pixel_embedding}
S.~Kong and C.~Fowlkes.
\newblock Recurrent pixel embedding for instance grouping.
\newblock {\em arXiv}, 2017.

\bibitem{kong2016photo}
S.~Kong, X.~Shen, Z.~Lin, R.~Mech, and C.~Fowlkes.
\newblock Photo aesthetics ranking network with attributes and content
  adaptation.
\newblock In {\em European Conference on Computer Vision}, pages 662--679.
  Springer, 2016.

\bibitem{koltun2011efficient}
P.~Krahenbuhl and V.~Koltun.
\newblock Efficient inference in fully connected crfs with gaussian edge
  potentials.
\newblock {\em NIPS}, 2011.

\bibitem{ladicky2014pulling}
L.~Ladicky, J.~Shi, and M.~Pollefeys.
\newblock Pulling things out of perspective.
\newblock In {\em Proceedings of the IEEE Conference on Computer Vision and
  Pattern Recognition}, 2014.

\bibitem{laina2016deeper}
I.~Laina, C.~Rupprecht, V.~Belagiannis, F.~Tombari, and N.~Navab.
\newblock Deeper depth prediction with fully convolutional residual networks.
\newblock In {\em 3D Vision (3DV), 2016 Fourth International Conference on},
  pages 239--248. IEEE, 2016.

\bibitem{lee2009geometric}
D.~C. Lee, M.~Hebert, and T.~Kanade.
\newblock Geometric reasoning for single image structure recovery.
\newblock In {\em Computer Vision and Pattern Recognition, 2009. CVPR 2009.
  IEEE Conference on}, pages 2136--2143. IEEE, 2009.

\bibitem{li2016iterative}
K.~Li, B.~Hariharan, and J.~Malik.
\newblock Iterative instance segmentation.
\newblock In {\em Proceedings of the IEEE Conference on Computer Vision and
  Pattern Recognition}, 2016.

\bibitem{lin2016refinenet}
G.~Lin, A.~Milan, C.~Shen, and I.~Reid.
\newblock Refinenet: Multi-path refinement networks with identity mappings for
  high-resolution semantic segmentation.
\newblock In {\em Proceedings of the IEEE Conference on Computer Vision and
  Pattern Recognition}, 2017.

\bibitem{lin2016efficient}
G.~Lin, C.~Shen, A.~van~den Hengel, and I.~Reid.
\newblock Efficient piecewise training of deep structured models for semantic
  segmentation.
\newblock In {\em Proceedings of the IEEE Conference on Computer Vision and
  Pattern Recognition}, 2016.

\bibitem{lindeberg1998feature}
T.~Lindeberg.
\newblock Feature detection with automatic scale selection.
\newblock {\em International journal of computer vision}, 30(2):79--116, 1998.

\bibitem{liu2015deep}
F.~Liu, C.~Shen, and G.~Lin.
\newblock Deep convolutional neural fields for depth estimation from a single
  image.
\newblock In {\em Proceedings of the IEEE Conference on Computer Vision and
  Pattern Recognition}, 2015.

\bibitem{long2015fully}
J.~Long, E.~Shelhamer, and T.~Darrell.
\newblock Fully convolutional networks for semantic segmentation.
\newblock In {\em Proceedings of the IEEE Conference on Computer Vision and
  Pattern Recognition}, 2015.

\bibitem{luo2017unsupervised}
Z.~Luo, B.~Peng, D.-A. Huang, A.~Alahi, and L.~Fei-Fei.
\newblock Unsupervised learning of long-term motion dynamics for videos.
\newblock {\em arXiv:1701.01821}, 2017.

\bibitem{romera2016recurrent}
B.~Romera-Paredes and P.~H.~S. Torr.
\newblock Recurrent instance segmentation.
\newblock In {\em ECCV}, 2016.

\bibitem{saxena2005learning}
A.~Saxena, S.~H. Chung, and A.~Y. Ng.
\newblock Learning depth from single monocular images.
\newblock In {\em NIPS}, 2005.

\bibitem{silberman2012indoor}
N.~Silberman, D.~Hoiem, P.~Kohli, and R.~Fergus.
\newblock Indoor segmentation and support inference from rgbd images.
\newblock {\em ECCV}, 2012.

\bibitem{song2015sun}
S.~Song, S.~P. Lichtenberg, and J.~Xiao.
\newblock Sun rgb-d: A rgb-d scene understanding benchmark suite.
\newblock In {\em Proceedings of the IEEE Conference on Computer Vision and
  Pattern Recognition}, 2015.

\bibitem{vedaldi2015matconvnet}
A.~Vedaldi and K.~Lenc.
\newblock Matconvnet: Convolutional neural networks for matlab.
\newblock In {\em Proceedings of the 23rd ACM international conference on
  Multimedia}, 2015.

\bibitem{zhao2016pyramid}
H.~Zhao, J.~Shi, X.~Qi, X.~Wang, and J.~Jia.
\newblock Pyramid scene parsing network.
\newblock In {\em Proceedings of the IEEE Conference on Computer Vision and
  Pattern Recognition}, 2017.

\bibitem{zheng2015conditional}
S.~Zheng, S.~Jayasumana, B.~Romera-Paredes, V.~Vineet, Z.~Su, D.~Du, C.~Huang,
  and P.~H. Torr.
\newblock Conditional random fields as recurrent neural networks.
\newblock In {\em Proceedings of the IEEE International Conference on Computer
  Vision}, pages 1529--1537, 2015.

\end{thebibliography}
\bibliographystyle{ieee} % abbr
}

\newpage
\newpage
\clearpage\mbox{}Page \thepage\ of the manuscript.

\section{ Analysis of Depth-aware Gating Module}
\label{sec:analysis-depth-aware-module}

\begin{figure*}[t]
\centering
   \includegraphics[width=1\linewidth]{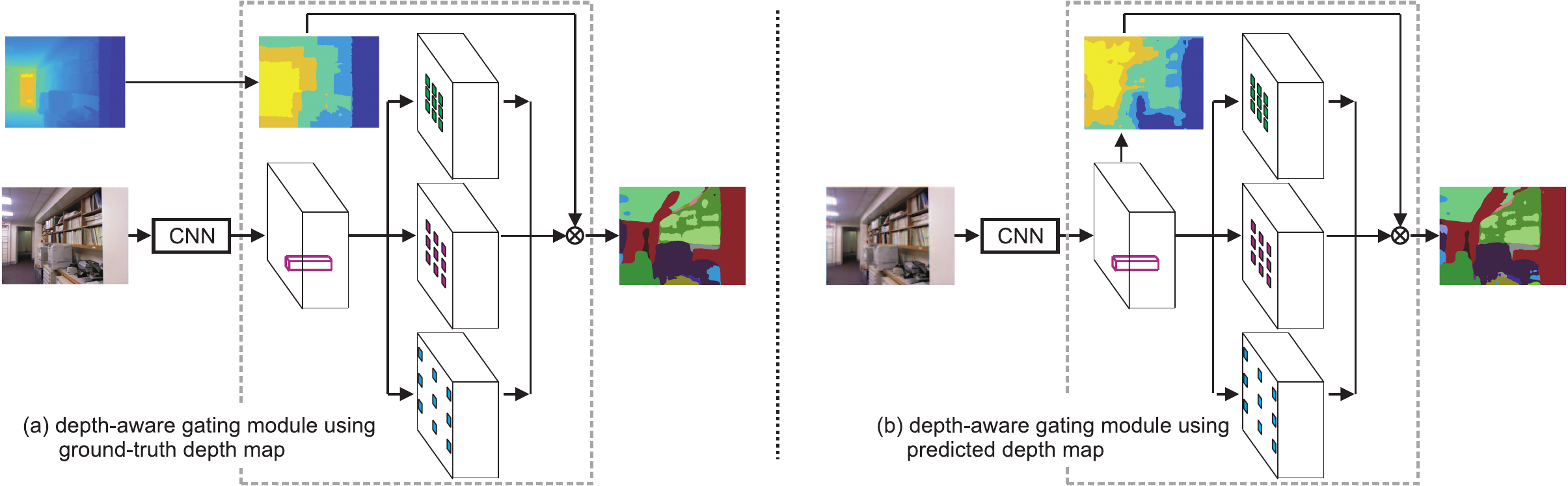}
   \caption{(a) Depth-aware gating module using the ground-truth depth map, and (b) depth-aware gating module using the predicted depth map.
   The grids within the feature map blocks distinguish different pooling field sizes.
   Here we depict three different pooling window sizes while in our actual experiments we quantize the depth map into five scale bins.
   }
\label{fig:depthGatingModule_appendix}
\end{figure*}

In this section, we analyze the proposed depth-aware gating module with detailed results in Table~\ref{tab:analysis-depth-aware-module}.
We perform the ablation study on the Cityscapes dataset~\cite{cordts2016cityscapes}.
Specifically,
we train the following models in order (except the fourth model which learns attention to gate).
\begin{enumerate}
  \item ``baseline''  is our DeepLab-like baseline model by training two
  convolutional (with $3\times 3$ kernels) layers above the ResNet101 backbone.

 \item ``tied, avg.'' is the model we train based on ``baseline'' by using the
  same $3\times 3$ kernel but different dilate rates equal to $\{1,2,4,8,16\}$,
  respectively. So there are five branches and each of them has the same
  kernels which are tied to make processing scale-invariant. We average the
  feature maps for the final output prior to classification.

  \item ``gt-depth, tied, gating'' is the model using the ground-truth depth
  map to select the branch; the pooling window size is determined
  according to the ground-truth depth value.

  \item ``gt-depth, untied, gating'' is the model based on ``gt-depth, tied,
  gating'' by unleashing the tied kernels in the five branches. These untied
  kernels improve the flexibility and representation power of the network.
  Figure~\ref{fig:depthGatingModule_appendix} (a) depicts this model.

  \item ``attention, untied, gating'' is an independent model to the previous ones that is trained
  without depth supervision where the gating signal acts as a generic attentional signal that modulates spatially adaptive pooling.
  Specifically,
  we train an attention branch to produce the soft weight mask after softmax to gate the features from multiple pooling at different scales.
  We also adopt untied weights for the scale-specific pooling branches.
  The architecture is similar to what depicted in Figure~\ref{fig:depthGatingModule_appendix} (b),
  but without depth supervision.

  \item ``pred-depth, untied, gating'' is our final model in which we learn a
  quantized depth predictor to gate the five branches.  This model determines
  the size of pooling window based on its predicted depth map.
  Figure~\ref{fig:depthGatingModule_appendix} (b) shows the architecture of this model.
\end{enumerate}

Through Table~\ref{tab:analysis-depth-aware-module},
we can see that increasing the dilate rate with our model ``tied, avg.'' improves the performance noticeably.
This is consistent with the observation in~\cite{chen2016deeplab},
in which the large view-of-field version of DeepLab performs better.
The benefit can be explained by the large dilation rate increasing the size of
the receptive field, allowing more contextual information to be captured at
higher levels of the network.
With the gating mechanism, either using ground-truth depth map or the predicted one,
the performance is improved further over non-adaptive pooling.
The depth-aware gating module helps determine the pooling window size wisely,
which is better than averaging all branches equally as in our ``tied, avg.'' model and DeepLab.
Moreover,
by unleashing the tied kernels,
the ``gt-depth untied, gating'' improves over ``gt-depth, tied, gating'' remarkably.
We conjecture that this is because the untied kernels provide more flexibility
to distinguish features at different scales and allow selection of the appropriate
non-invariant features from lower in the network.
Interestingly,
the attention-gating model performs well and using the predicted depth map achieves the best among all these compared models.
We attribute this to three reasons.
Firstly, the predicted depth is smooth without holes or invalid entries.
%so is the attention map.
When using ground-truth depth on Cityscapes dataset,
we assign equal weight on the missing entries so that the gating actually averages the information at different scales.
This average pooling might be harmful in some cases such as very small object at distance.
This can be taken as complementary evidence that the blindly averaging all branches
achieves inferior performance to using the depth-aware gating.
Secondly,
the predicted depth maps have some object-aware pattern structure, which might
be helpful for segmentation.
From the visualization shown later in Figure~\ref{fig:appendix_Cityscapes},
we can observe such patterns, e.g. for cars.
%Whereas the attention map captures distance transform around segment boundaries, as shown also in Figure~\ref{fig:appendix_Cityscapes}, meaning that the model tries to avoid pooling across segments.
%This is understandably a desired property for pixel-wise prediction problems.
Thirdly,
the depth prediction branch, as well as the attention branch,
generally increases the representation power and
flexibility of the whole model; this can be beneficial for segmentation.

\begin{sidewaystable*}%[th]
\small % \footnotesize
\centering
\caption{Result of different depth-aware gating module deployments on Cityscapes dataset. IoU is short for intersection over union averaged over all classes,
and nIoU is the weighted IoU through the pre-defined class weights provided by the benchmark. }
%\vspace{-3mm}
\begin{tabular}{l  c c c    c c c   c c c   c c c   c c c   c c c  }
\hline
         &   \multicolumn{2}{c} {baseline} & &  \multicolumn{2}{c} {tied, avg.} & &\multicolumn{2}{c} {gt-depth, tied, gating} & &\multicolumn{2}{c} {gt-depth, untied, gating}  &
         &\multicolumn{2}{c} {attention, untied, gating} &
         &\multicolumn{2}{c} {pred-depth, untied, gating} \\
    \cmidrule(r){2-3} \cmidrule(r){5-6} \cmidrule(r){8-9} \cmidrule(r){11-12} \cmidrule(r){14-15}
       &   IoU      & nIoU &&   IoU      & nIoU &&   IoU      & nIoU & &  IoU      & nIoU &&   IoU      & nIoU      \\
\hline
\textbf{Score Avg. }   & \textbf{0.738}&    \textbf{0.547}   & &  \textbf{0.747}    &    \textbf{0.554}& &  \textbf{0.748}&  \textbf{0.556}& &  \textbf{0.753}&    \textbf{0.561}& &
 \textbf{0.754}&    \textbf{0.558}& &  \textbf{0.759}&    \textbf{0.571}\\
\hline
road          & 0.980&      --   & &  0.981    &      --& &   0.981&      -- &   &0.982&      --& & 0.982&      --& &    0.982&      --      \\
sidewalk      & 0.849&      --   & &  0.847    &      --& &   0.849&      -- &   &0.982&      --& & 0.853&      --& &    0.857&      --      \\
building      & 0.916&      --   & &  0.917    &      --& &   0.918&      -- &   &0.919&      --& & 0.923&      --& &    0.920&      --      \\
wall          & 0.475&      --   & &  0.499    &      --& &   0.506&      -- &   &0.511&      --& & 0.527&      --& &    0.512&      --      \\
fence         & 0.596&      --   & &  0.605    &      --& &   0.605&      -- &   &0.611&      --& & 0.618&      --& &    0.614&      --      \\
pole          & 0.598&      --   & &  0.599    &      --& &   0.604&      -- &   &0.616&      --& & 0.615&      --& &    0.624&      --      \\
traffic light & 0.684&      --   & &  0.674    &      --& &   0.678&      -- &   &0.692&      --& & 0.689&      --& &    0.699&      --      \\
traffic sign  & 0.780&      --   & &  0.776    &      --& &   0.775&      -- &   &0.782&      --& & 0.783&      --& &    0.790&      --      \\
vegetation    & 0.918&      --   & &  0.917    &      --& &   0.918&      -- &   &0.920&      --& & 0.920&      --& &    0.922&      --      \\
terrain       & 0.619&      --   & &  0.620    &      --& &   0.627&      -- &   &0.632&      --& & 0.625&      --& &    0.638&      --      \\
sky           & 0.941&      --   & &  0.937    &      --& &   0.940&      -- &   &0.942&      --& & 0.943&      --& &    0.944&      --      \\
person        & 0.803&    0.635   & &  0.803    &    0.631& &   0.804&    0.639 &   &0.808&    0.648& & 0.804&  0.641& & 0.814&    0.659      \\
rider         & 0.594&    0.448   & &  0.595    &    0.462& &   0.602&    0.460 &   &0.612&    0.461& & 0.602&  0.443& & 0.616&    0.473      \\
car           & 0.939&    0.859   & &  0.942    &    0.854& &   0.942&    0.863 &   &0.942&    0.867& & 0.943&  0.862& & 0.944&    0.871      \\
truck         & 0.631&    0.398   & &  0.666    &    0.421& &   0.679&    0.407 &   &0.679&    0.417& & 0.666&  0.424& & 0.674&    0.421      \\
bus           & 0.759&    0.595   & &  0.802    &    0.607& &   0.787&    0.612 &   &0.786&    0.609& & 0.798&  0.602& & 0.799&    0.615      \\
train         & 0.621&    0.467   & &  0.683    &    0.494& &   0.656&    0.489 &   &0.655&    0.487& & 0.684&  0.508& & 0.687&    0.507      \\
motorcycle    & 0.562&    0.396   & &  0.587    &    0.387& &   0.591&    0.398 &   &0.602&    0.410& & 0.583&  0.407& & 0.610&    0.425      \\
bicycle       & 0.755&    0.582   & &  0.747    &    0.387& &   0.753&    0.575 &   &0.761&    0.586& & 0.760&  0.575& & 0.765&    0.594      \\
\hline
\end{tabular}
\label{tab:analysis-depth-aware-module}
\end{sidewaystable*}

\section{ Results on the SUN-RGBD dataset}
\label{sec:SUNRGBD}

\begin{figure*}[t]
\centering
   \includegraphics[width=1\linewidth]{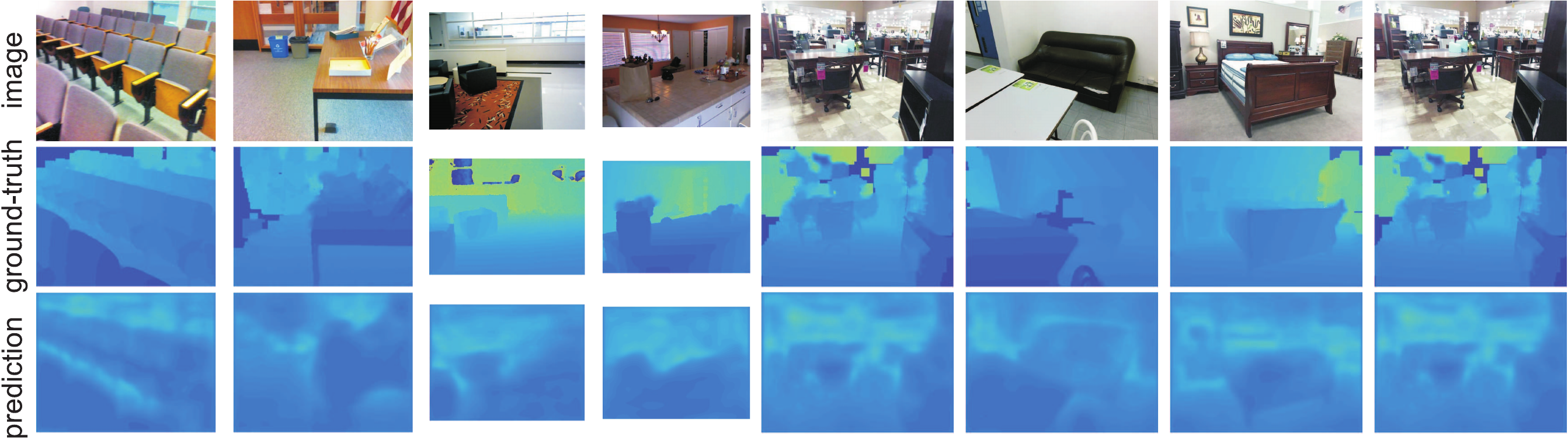}
   \caption{Visualization of images from SUN-RGBD dataset and their ground-truth depth and our predicted depth on the three rows, respectively.
    We scale all the depth maps into a fixed range of $[0, 10^5]$.
    In this sense, the color of the depth maps directly reflect the absolute physical depth.
    Note that there are unnatural regions in the ground-truth depth maps, which have been refined by the algorithm in~\cite{song2015sun}.
    Visually, these refined region do not always make sense and are incorrect depth completions.
    In contrast, our monocular predictions are quite smooth.
    }
\label{fig:depth_SUN-RGBD}
\end{figure*}

\begin{figure*}[t]
\centering
   \includegraphics[width=1\linewidth]{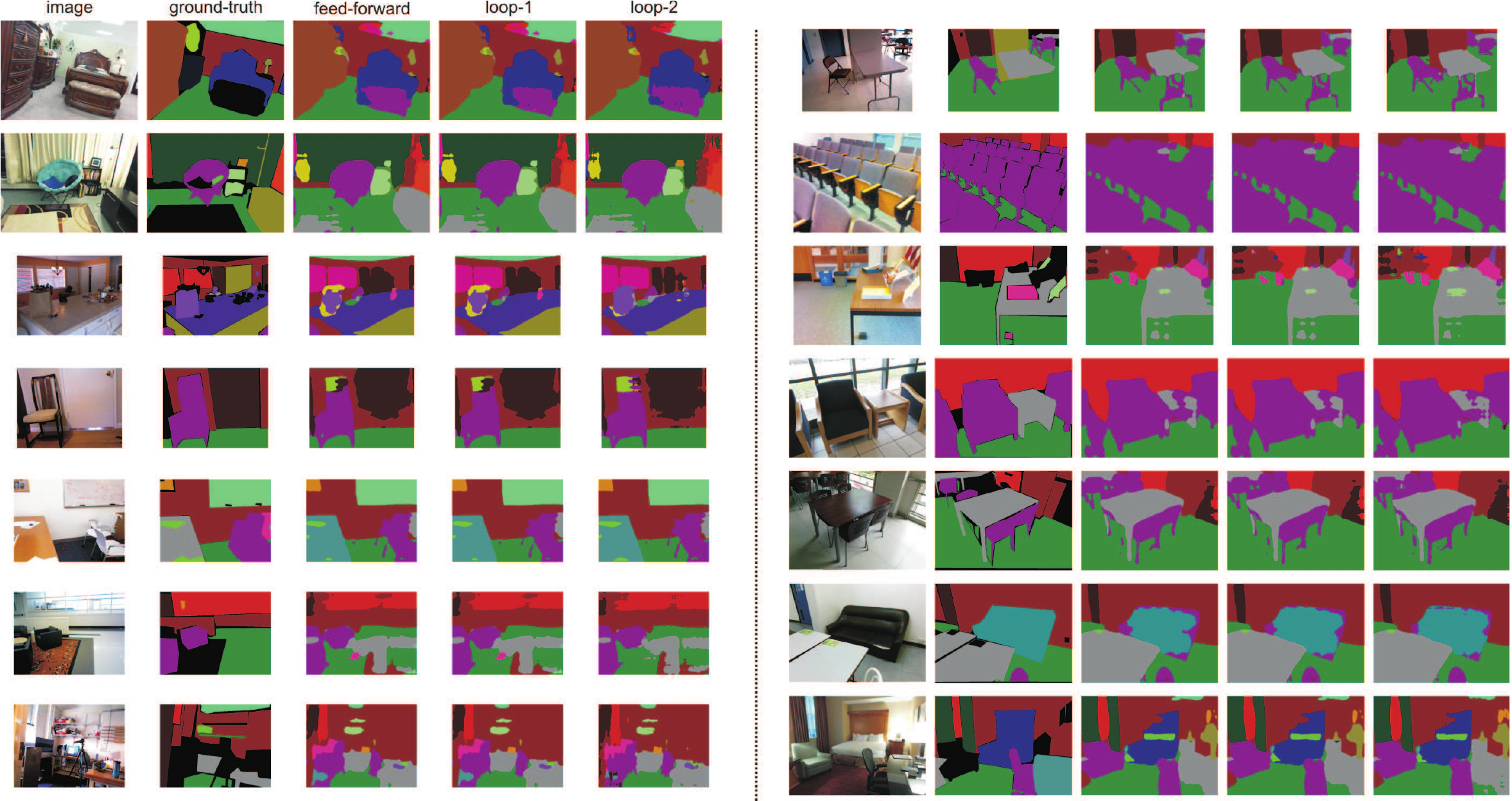}
   \caption{Visualization of the output on SUN-RGBD dataset.
   We randomly show fourteen images from validation set with their segmentation output from both feed-forward pathway and recurrent loops.
   In the ground-truth segmentation annotation,
   we can see that there are many regions (with black color) not annotated.}
\label{fig:visualization_SUN-RGBD}
\end{figure*}

In Figure~\ref{fig:depth_SUN-RGBD},
we show the depth prediction results of several images randomly picked from the test set of SUN-RGBD dataset.
Note that the there are unnatural regions in the ground-truth depth maps, which are the result of refined depth
completion by the algorithm in~\cite{song2015sun}.
Visually, these regions do not always make sense and constitute bad depth completions.
In contrast,
our predicted depth maps are much smoother.
We also evaluate our depth prediction on
SUN-RGBD dataset, and achieve 0.754, 0.899 and 0.961 by the three threshold
metrics respectively.  As SUN-RGBD is an extension of NYU-depth-v2 dataset, it
has similar data statistics resulting in similar prediction performance.

In Figure~\ref{fig:visualization_SUN-RGBD},
we randomly show fourteen images and their segmentation results at loops of the recurrent refining module.
Visually,
we can see that the our recurrent module refines the segmentation result in the loops.

\section{  Visualization on Large Perspective Images}
\label{sec:perspective}

\begin{figure*}[t]
\centering
   \includegraphics[width=1\linewidth]{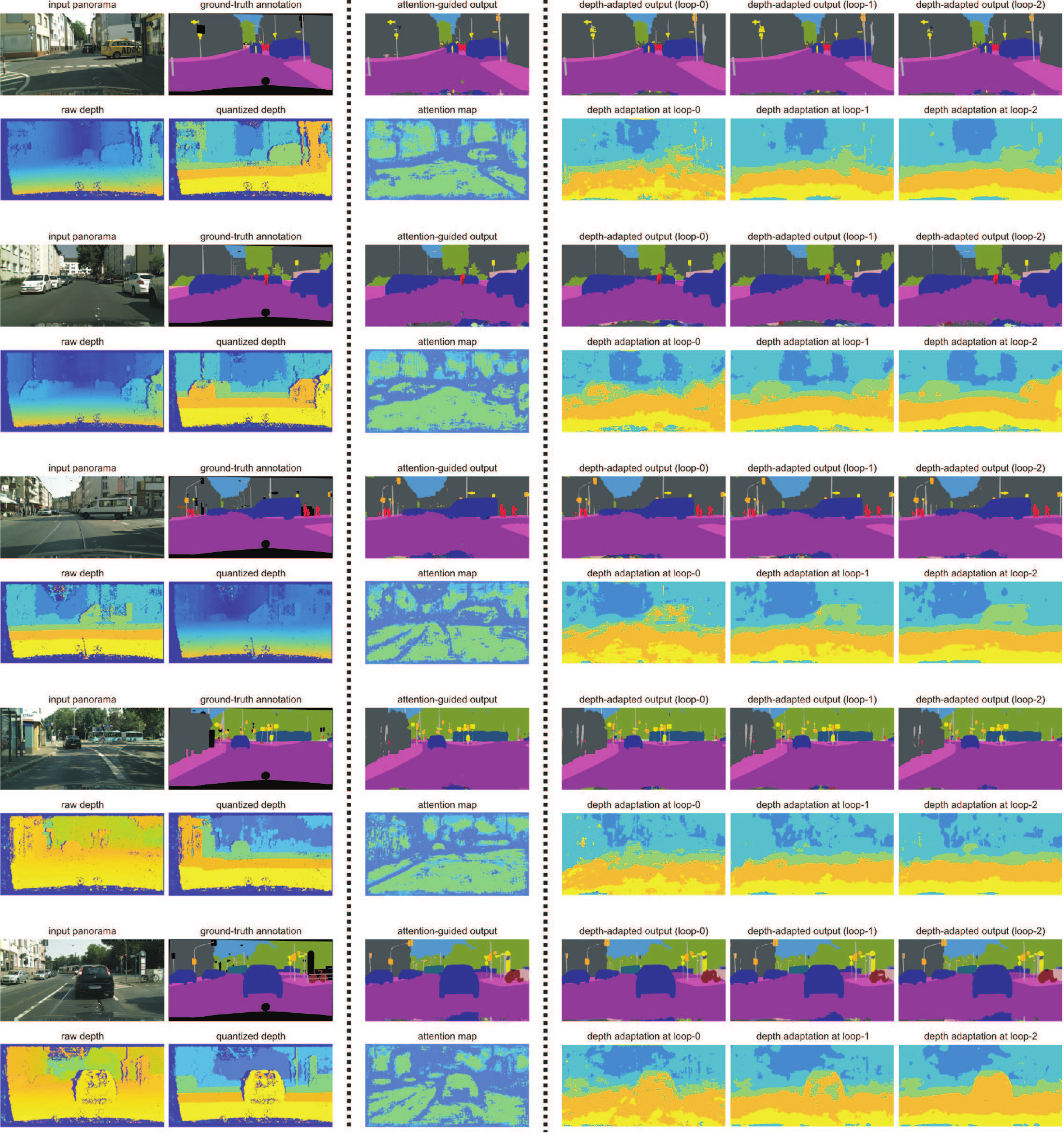}
   \caption{Visualization of the results on Cityscapes dataset.
   For five random images from the validation set,
   we show the input perspective street scene photos,
   ground-truth annotation,
   raw disparity and the five-scale quantized depth map in the leftmost two columns.
   Then,
   we show the segmentation prediction and the attention map using our unsupervised attentional mechanism in the third column.
   In the rest three columns,
   we show the output of our depth-aware adaptation within recurrent refinement,
   from loop-0 to loop-2.
   Note that the more yellowish the color is,
   the closer the object is to camera and the finer scale of the feature maps the model adopts to process.
   From the visualization,
   we can see 1) the attention map helps the model avoid pooling across semantic segments;
   2) the depth-adaptation in the recurrent refinement loops gradually captures objects like the cars,
   we attribute this to to the top-down signal from previous loops.
   }
\label{fig:appendix_Cityscapes}
\end{figure*}

\begin{figure*}[t]
\centering
   \includegraphics[width=1\linewidth]{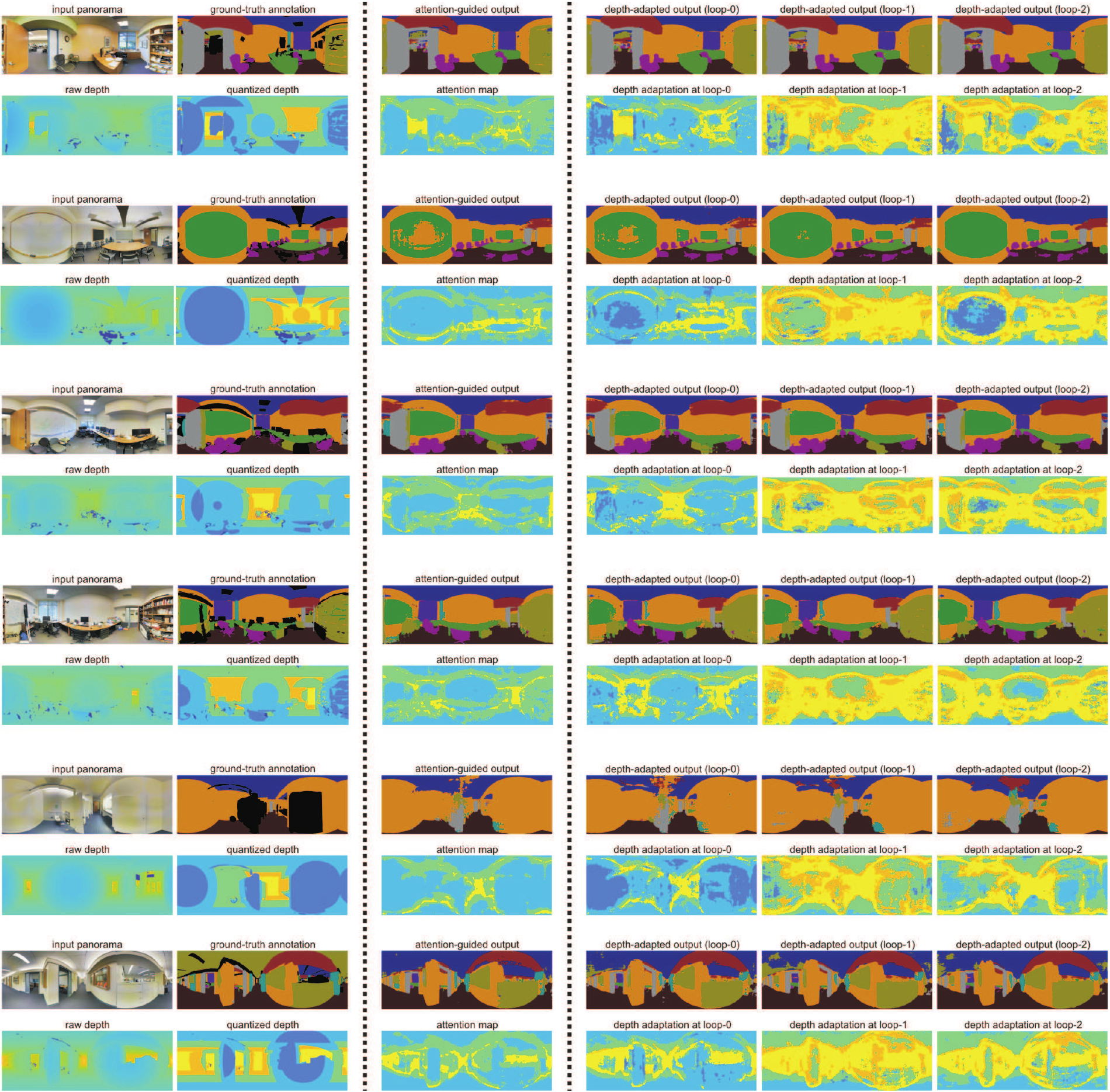}
   \caption{Visualization of the results on Stanford-2D-3D dataset.
   For six random images from the validation set,
   we show the input panorama, ground-truth annotation, raw depth map and the five-scale quantized depth map in the leftmost two columns.
   Then,
   we show the segmentation prediction and the attention map using our unsupervised attentional mechanism in the third column.
   In the rest three columns,
   we show the output of our depth-aware adaptation within recurrent refinement,
   from loop-0 to loop-2.
   Note that the more yellowish the color is,
   the further away the object is to camera and the finer scale of the feature maps the model adopts to process.
   From the visualization,
   we can see 1) the attention map helps the model avoid pooling across semantic segments;
   2) the depth-adaptation in the recurrent refinement loops fulfill coarse-to-fine processing as smaller receptive fields are used, due to the top-down signal from previous loops.
   }
\label{fig:appendix_Stanford2D3D}
\end{figure*}

In Figure~\ref{fig:appendix_Cityscapes} and \ref{fig:appendix_Stanford2D3D},
we visualize more results on Cityscapes and Stanford-2D-3D datasets, respectively.
First,
we show the segmentation prediction and the attention map after training with the unsupervised attentional mechanism in the third column.
We can see the attention map appears to encode the distance from object boundaries.
We hypothesize this selection mechanism serves to avoid pooling features across different semantic segments while still
utilizing large pooling regions within each region.
This is understandable and desirable in practice,
as per-pixel feature vectors have different feature statistics for different categories.
Then,
we compare the segmentation results and depth estimate for adaptation in the recurrent refinement loops (last three columns in Figure~\ref{fig:appendix_Cityscapes} and \ref{fig:appendix_Stanford2D3D}).
We notice that the depth estimate for adaptation changes remarkably in the loop (the depth module is fine-tuned using the segmentation loss only in training).
While the depth estimate captures some object shapes in Cityscapes (e.g. car),
it becomes more noticeable that the depth prediction helps the model perform coarse-to-fine refinement in the loop by using smaller receptive fields in Stanford-2D-3D dataset.
We conjecture that this is owing to the top-down signal from the depth estimate at the previous loop.
The recurrent refinement module also fills the holes in large areas, like light reflection regions on the car in street scene (Cityscapes) and white board in the second image (row 3 and 4) of panoramic photos (Stanford-2D-3D).

\end{document}